\documentclass[fleqn,10pt]{wlscirep}
\PassOptionsToPackage{table,xcdraw}{xcolor}
\usepackage{placeins}
\usepackage{subcaption}
\usepackage{cleveref}
\usepackage{soul}
\usepackage{booktabs}
\usepackage{amsmath}
\usepackage{graphicx}
\usepackage[utf8]{inputenc}
\usepackage{listings}
\usepackage{algorithm2e}
\usepackage{graphicx}
\usepackage{multirow}
\usepackage{tikz}
\usepackage{adjustbox}
\usepackage{pdflscape}
\usepackage{url}
\usepackage{float}
\usepackage{array}
\usepackage{multirow}
\usepackage{tcolorbox}
\usepackage{upquote}
\usepackage{colortbl}
\usepackage[backgroundcolor=softPeach]{todonotes}
\usepackage{tabularx}
\usepackage{geometry}
 
\definecolor{codegreen}{rgb}{0,0.6,0}
\definecolor{codegray}{rgb}{0.5,0.5,0.5}
\definecolor{codepurple}{rgb}{0.58,0,0.82}
\definecolor{backcolour}{rgb}{0.95,0.95,0.92}
\definecolor{softPeach}{RGB}{255, 200, 180}

\lstdefinestyle{mystyle}{ backgroundcolor=\color{backcolour},    commentstyle=\color{codegreen}, keywordstyle=\color{magenta}, numberstyle=\tiny\color{codegray}, stringstyle=\color{codepurple}, basicstyle=\ttfamily\footnotesize,
    breakatwhitespace=false,  
    breaklines=true,          
    captionpos=b,             
    keepspaces=true,          
    numbers=left,             
    numbersep=5pt,            
    showspaces=false,
    showstringspaces=false,
    showtabs=false,
    tabsize=2
}

\title{Endoscopic Video Dataset For Automated Surgical Timeline Segmentation Using Computational Artificial Intelligence Techniques}
\title{Data Descriptor Title (110 characters maximum, inc. spaces)}
\title{Endoscopic Video Dataset: Advancing Computational AI for Surgical Scene Analysis in Colorectal Cancer Operations}
\title{Hierarchical Surgical Timeline Segmentation Dataset of Surgical Videos: automated indexing of Video-Based Assessments}
\title{EndoSTG-Dataset: Hierarchical Surgical Timeline Segmentation and Indexing of Transanal Endoscopic Microsurgery Videos}
\title{TEMS-STG: Transanal Endoscopic Microsurgery Videos Dataset for Hierarchical Surgical Timeline Generation and Automatic Indexing of Video-Based Assessments}
\title{DATEM: A Novel Densely Annotated TEMS Dataset for Indexing Large Surgical Videos via Timeline Segmentation}
\title{TEMSET-50K: A Novel Densely Annotated Dataset for Indexing Large Surgical Videos via Spatiotemporal Timeline Segmentation}
\title{TEMSET-24K: Densely Annotated Dataset for Indexing Multipart Endoscopic Videos using Surgical Timeline Segmentation}

\author[1,*,$\dag$]{Muhammad Bilal}
\author[1]{Mahmood Alam}
\author[2]{Deepa Bapu}
\author[2]{Stephan Korsgen}
\author[2]{Neeraj Lal}
\author[2,3]{Simon Bach}
\author[3]{Amir M Hajivanand}
\author[2]{Muhammed Ali}
\author[4]{Kamran Soomro}
\author[5]{Iqbal Qasim}
\author[4]{Paweł Capik}
\author[6]{Aslam Khan}
\author[4]{Zaheer Khan}
\author[7]{Hunaid Vohra}
\author[7]{Massimo Caputo}
\author[2,3]{Andrew Beggs}
\author[8]{Adnan Qayyum}
\author[7]{Junaid Qadir}
\author[1,2,3$\dag$]{Shazad Ashraf}
\affil[1]{Birmingham City University, United Kingdom}
\affil[2]{University Hospitals Birmingham, Birmingham, United Kingdom}
\affil[3]{University of Birmingham, Birmingham, United Kingdom}
\affil[4]{University of the West of England, Bristol, United Kingdom}
\affil[5]{University of Hertfordshire, United Kingdom}
\affil[6]{University of Bradford, United Kingdom}
\affil[7]{University of Bristol, United Kingdom}
\affil[8]{Information Technology University, Lahore, Pakistan}
\affil[9]{Qatar University, Doha, Qatar}

\affil[*]{corresponding author(s): Muhammad Bilal (muhammad.bilal@bcu.ac.uk); Shazad Ashraf (S.ashraf.2@bham.ac.uk)}

\affil[$\dag$]{these authors contributed equally to this work}

\begin{abstract}
Indexing endoscopic surgical videos is vital in surgical data science, forming the basis for systematic retrospective analysis and clinical performance evaluation. Despite its significance, current video analytics rely on manual indexing, a time-consuming process. Advances in computer vision, particularly deep learning, offer automation potential, yet progress is limited by the lack of publicly available, densely annotated surgical datasets.

\vspace{2mm}

To address this, we present TEMSET-24K, an open-source dataset comprising 24,306 trans-anal endoscopic microsurgery (TEMS) video micro-clips. Each clip is meticulously annotated by clinical experts using a novel hierarchical labeling taxonomy encompassing “phase, task, and action” triplets, capturing intricate surgical workflows. To validate this dataset, we benchmarked deep learning models, including transformer-based architectures. Our in silico evaluation demonstrates high accuracy (up to 0.99) and F1 scores (up to 0.99) for key phases like “Setup” and “Suturing.” The STALNet model, tested with ConvNeXt, ViT, and SWIN V2 encoders, consistently segmented well-represented phases. TEMSET-24K provides a critical benchmark, propelling state-of-the-art solutions in surgical data science.

\vspace{2mm}

\vspace{2mm}

\vspace{2mm}

\vspace{2mm}

\end{abstract}
\begin{document}

\flushbottom
\maketitle
\thispagestyle{empty}


\section*{Background \& Summary}
Over 300 million surgical procedures are performed worldwide annually \cite{weiser2015estimate}. While surgery is a crucial healthcare intervention, it also carries significant risks, with surgical complications currently ranking as the third leading cause of global mortality \cite{nepogodiev2019global}. Surgical adverse events result in major quality-of-life (QoL) issues for patients, and methods that critically evaluate intra-operative events have significant potential to drive up surgical standards and reduce morbidity. A key method for enhancing surgical standards involves the use of high-resolution endoscopic surgical videos (ESV). These videos capture minimally invasive surgeries (MIS) with high-definition visual records at 60 frames per second, producing two simultaneous full HD streams. This results in over 50GB of data for a single uncompressed video, with even greater volumes as surgeries lengthen, or 4K resolution technology is adopted. This poses significant challenges when attempting to create adequate storage capacity in Secure Digital Environments (SDEs), such as those recently implemented by the National Health Service (NHS), UK \cite{SDE}. Despite the storage and energy costs, the value of ESV files in capturing surgical details is significant, especially at scale. Reduction in storage requirements without loss of vital information will inevitably lead to significant energy and cost savings in line with the plan to reduce the NHS Carbon Footprint to zero by 2040 \cite{Green}. 

\vspace{2mm}
Apart from the storage and management challenges, another major stumbling block hindering surgical scene understanding is the lack of richly annotated, comprehensive datasets. A meticulously assembled large dataset is invaluable for training machine learning models to recognise objects like instruments and anatomical structures in the surgical field of view and to understand procedural phases, tasks, and intra-operative actions. Such capabilities in scene synthesis, facilitated by automated algorithms, are vital for elucidating the intricacies of surgical workflows \cite{maier2017surgical} and evaluating surgeon performance \cite{reiley2011review}. This underscores the importance of developing high-quality representative ESV datasets in a SDE to advance surgical data science and create SOTA vision tools for clinical use.

\vspace{2mm}
Recent advancements in video-based analysis (VBA) using AI-driven computer vision techniques present substantial opportunities for enhancing surgical scene understanding through more scalable and robust methodologies \cite{goodman2021real}. Tailoring these VBA approaches specifically for ESV is crucial for demonstrating their efficacy in surgical data science and their potential application in real-world clinical settings. At the core of surgical scene understanding is the segmentation of surgical timelines, which involves analysing video sequences to categorise diverse surgical elements---ranging from phases and tasks to activities and adverse events. Unlike object segmentation, which focuses on image-level analysis, timeline segmentation operates at the video level, presenting a volumetric and ``moving object'' challenge far more complex than natural scenes of stationary objects. Moreover, the high similarity between different surgical phases, variability in surgeon styles, inconsistent labelling, ambiguous workflow transitions, and the scarcity of annotated training data exacerbate the complexity. These challenges hinder the development of reliable digital tools for practical and widespread clinical use.

\vspace{2mm}
Additionally, manually reviewing extensive ESV files is time-consuming and inefficient for human clinical experts. If done systematically, this can take up significant time that could be used for other clinical tasks. Consequently, creating digital tools capable of conducting comprehensive and accurate evaluations of ESV clips becomes essential to propel advancements in the field. This study aims to: (a) establish a systematic methodology for curating a high-quality, ``densely'' annotated ESV dataset, (b) assess the performance of cutting-edge video analysis models for surgical timeline segmentation, and (c) validate the most effective model for indexing ESV files to enhance search capabilities. This paper outlines strategies for transitioning from laboratory in-silico models to clinical applications, aiming to harness AI's potential to enhance interventional care to drive up surgical standards. To integrate SOTA methodologies in surgical data science, the whole pathway from video recording to annotation and analysis must be digitised. This will help improve and standardise overall surgical task performance evaluation. In summary, we make the following salient contributions:
\begin{enumerate}
\item We present timeline annotation taxonomy for TEMS procedures capturing five phases, 12 tasks, and 84 actions for end-to-end surgical timeline segmentation.
\item We put forward \texttt{TEMSET-24K}---a densely annotated TEMS dataset comprising 24,306 cut-outs, keyframes, and label files capturing timeline labels for the proposed labelling strategy, the nature of the surgical action (adverse/normal), remaining surgical time, etc.
\item We share our endoscopic video review (\texttt{EVR}\footnotemark{}\footnotetext{\url{https://github.com/bilalcodehub/evr}}) Python library with the surgical data science community to perform the necessary pre-processing required for curating and managing large multi-clip surgical video datasets in other specialities.
\item We implement and evaluate STALNet ESV analytics using state-of-the-art encoders, including \texttt{ConvNeXt}, \texttt{ViT}, and \texttt{SWIN V2}, for surgical timeline segmentation to showcase the efficacy of the proposed taxonomy and benchmark the curated dataset for indexing surgical videos.
\end{enumerate}

\subsection*{Related Work}
This section discusses prior work related to timeline analysis in surgical videos and state-of-the-art VBA methods, highlighting the potential for integrating timeline recognition with VBA to enhance the performance and generalisability of surgical data science solutions.

\subsubsection*{Timeline Analysis in Surgical Scenes}
Timeline analysis in surgical videos involves breaking down surgical procedures into distinct phases, tasks, and actions to provide a comprehensive understanding of the surgical workflow. Detailed workflow specifications capture all surgical nuances using phase/task/action triplets, which are essential for designing intelligent systems in the clinical operating room. These systems can provide context-aware decision support, monitor and optimise surgical operations, and offer early alerts for potential deviations and anomalies \cite{padoy2019machine, huaulme2020offline}.

\vspace{2mm}
Numerous studies have focused on surgical workflow analysis to identify missing activities in distinct phases, ensuring that surgeons complete necessary tasks before moving to the next phase \cite{kadkhodamohammadi2021towards}. Techniques for identifying surgical phases include data from sensors on tool tracking systems \cite{holden2014feasibility}, binary signals from instrument usage \cite{padoy2012statistical}, and surgical robots \cite{lin2005automatic}. However, obtaining these signals typically requires additional hardware or time-consuming manual annotation, which could increase the workload associated with the surgical process \cite{dergachyova2016automatic}.

\vspace{2mm}
Recent studies have focused on deriving the workflow solely from routinely collected endoscopic videos during surgery \cite{jin2017sv}. Automatic workflow recognition from surgical videos eliminates the need for additional equipment \cite{blum2010modeling}. Notable studies include the development of EndoNet, a convolutional neural network (CNN) architecture designed to recognise surgical phases using only visual information from cholecystectomy procedures \cite{twinanda2016endonet}. Other studies have employed temporal CNN models and transformer-based models for phase recognition in surgical activities \cite{ramesh2021multi, gao2021trans}. For instance, Funke et al. proposed a temporal model, TUNeS, which integrates self-attention into a convolutional U-Net architecture to enhance surgical phase recognition \cite{funke2023tunes}.

\subsubsection*{Emergence of Video-Based Analytics}

VBA involves meticulously breaking down and examining video content to extract important insights and intra-operative key events, transforming visual streams into semantically meaningful representations that can be easily analysed at scale \cite{huber2020video, liu2009encyclopedia}. Understanding surgical scenes requires consideration of the temporal dimension, making VBA crucial for providing an accurate understanding of surgical processes by examining both spatial and temporal features \cite{feldman2020sages}.

\vspace{2mm}
Real-time VBA can significantly enhance surgical care, particularly for minimally invasive techniques, by providing context-aware intra-operative decision support using AI models that swiftly and accurately extract knowledge from real-time video data. This situational guidance can improve surgical outcomes by aiding in applications such as calculating surgery duration, recording important events, assessing surgical skills, and providing intra-operative assistance \cite{vercauteren2019cai4cai, nwoye2020recognition, mascagni2022computer}. However, timeline labels in most research often lack the detail required for realistic clinical tasks, providing only coarse-grained information that fails to encompass surgical phases, tasks, and discrete actions needed for objective assessment and benchmarking of surgical performance \cite{lewandrowski2020regional, richards2015national}.

\vspace{2mm}
Additionally, deep learning methods used for surgical timeline segmentation often require large amounts of annotated data, which is rarely available \cite{paysan2021self}. To address this, Valderrama et al. introduced the PSI-AVA Dataset, which provides comprehensive annotations for surgical scene understanding in prostatectomy videos, and proposed the TAPIR model for action recognition using transformers \cite{valderrama2022towards}. Similarly, Ayobi et al. presented the TAPIS model, also based on transformers, to facilitate multilevel comprehension of surgical activities, including long-term tasks and instrument segmentation and atomic action detection \cite{ayobi2024pixel}.

\vspace{2mm}
By combining insights from timeline analysis and VBA, our study aims to further advance the understanding and evaluation of surgical procedures through the development of high-quality, deeply annotated ESV datasets and the establishment of robust AI models for surgical timeline segmentation.

\section*{Methods}

\subsection*{Transanal Endoscopic Microsurgery (TEMS) Overview}
The dataset described in this paper comprises recordings of TEMS procedures performed on patients with early rectal cancer or large pre-cancerous polyps \cite{BACH202192}. During the TEMS procedure, an operating scope is inserted trans-anally into the rectum. This is stable and flexible platform that enables access from the anorectal junction to the most cephalic aspect of the rectum is 15cm from the anal verge (bottom of the anal canal). Most of the rectum can be reached with this TEMS scope. The surgeon adjusts the scope to reach and remove the tumour, manoeuvring it as needed. The procedure begins with a \texttt{setup} phase, which includes preparing the scope, instruments, and the surgical site. The rectum is inflated with carbon dioxide to a preset pressure, and faecal debris and fluid are removed with a suction device to obtain clear views. The main phase involves dissecting the tumour, removing the specimen, and closing the rectal wall defect. Surgeons use a \texttt{clockface} analogy to navigate the lesion site, facilitating precise removal. Dissection may be partial (mucosa and submucosa) or full thickness (deeper muscle tissues). 


    

\begin{figure}[!ht]
    \centering
    \begin{adjustbox}{width=\textwidth}
    \begin{tabular}{ccccc}
        \begin{subfigure}[b]{0.18\textwidth}
            \centering
            \includegraphics[height=3cm]{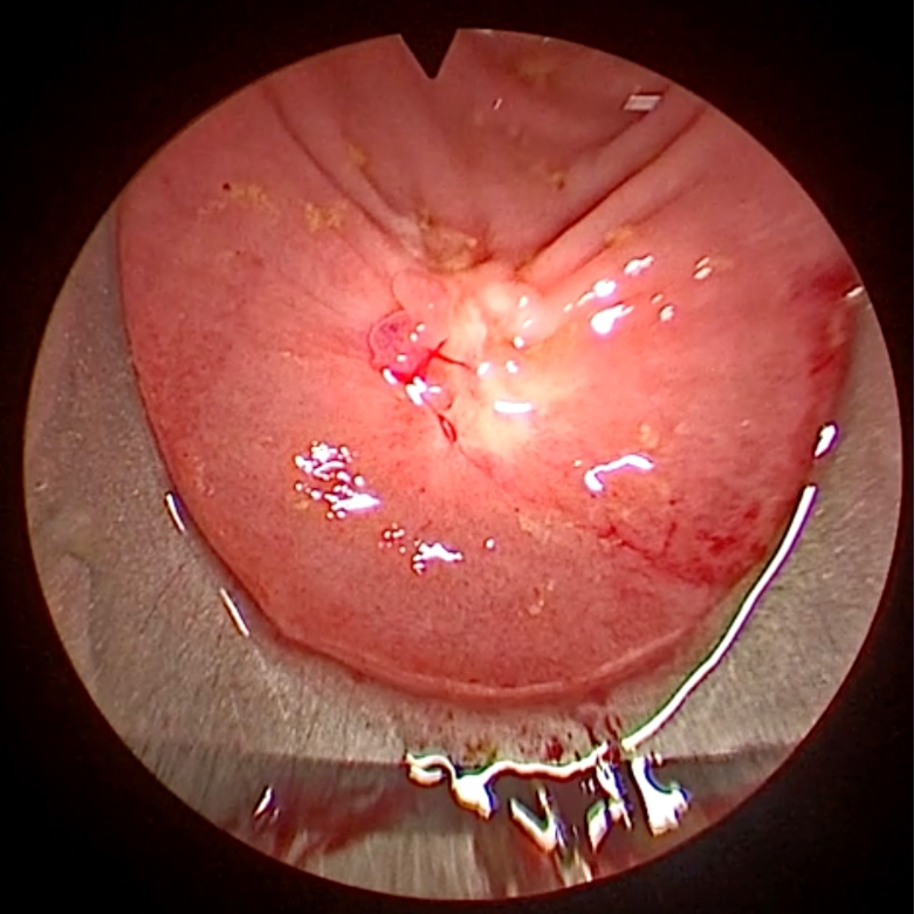}
            \caption*{A}
            \label{A}
        \end{subfigure} &
        \begin{subfigure}[b]{0.18\textwidth}
            \centering
            \includegraphics[height=3cm]{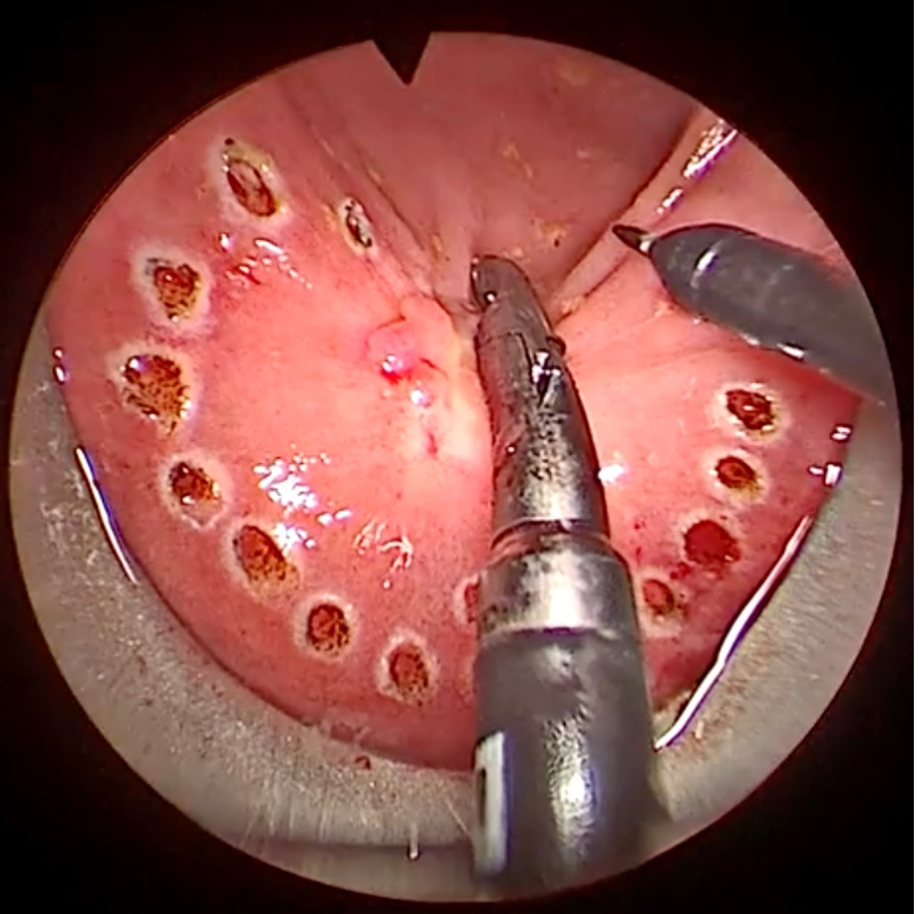}
            \caption*{B}
            \label{B}
        \end{subfigure} &
        \begin{subfigure}[b]{0.18\textwidth}
            \centering
            \includegraphics[height=3cm]{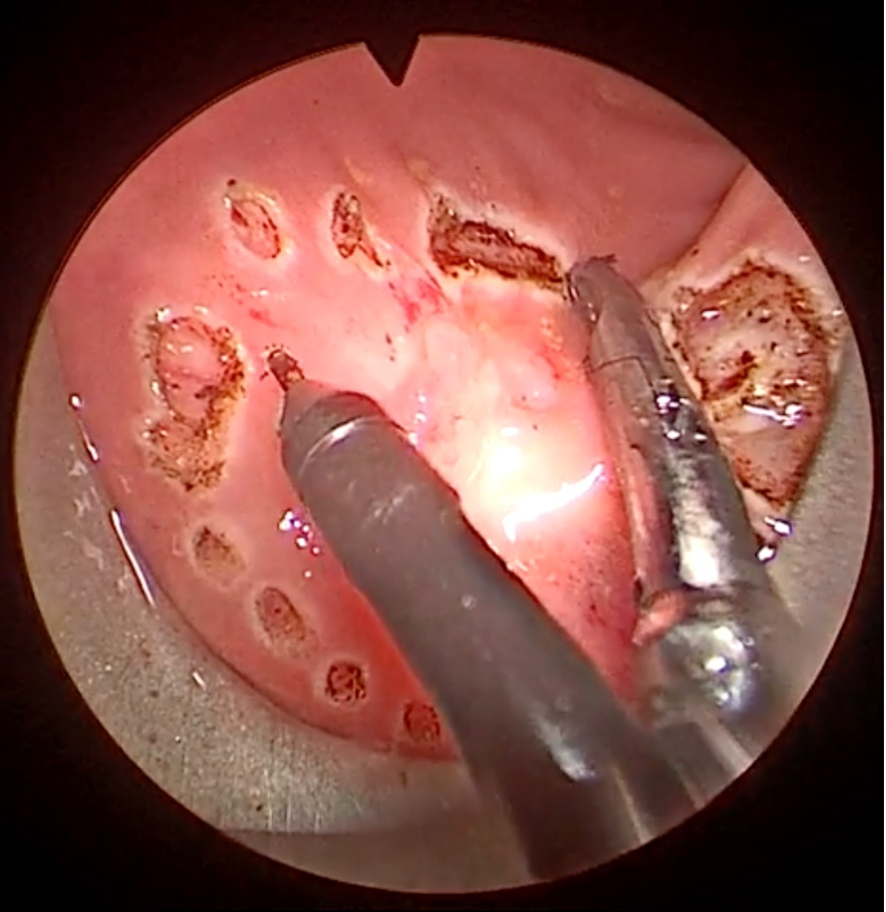}
            \caption*{C}
            \label{C}
        \end{subfigure} &
        \begin{subfigure}[b]{0.18\textwidth}
            \centering
            \includegraphics[height=3cm]{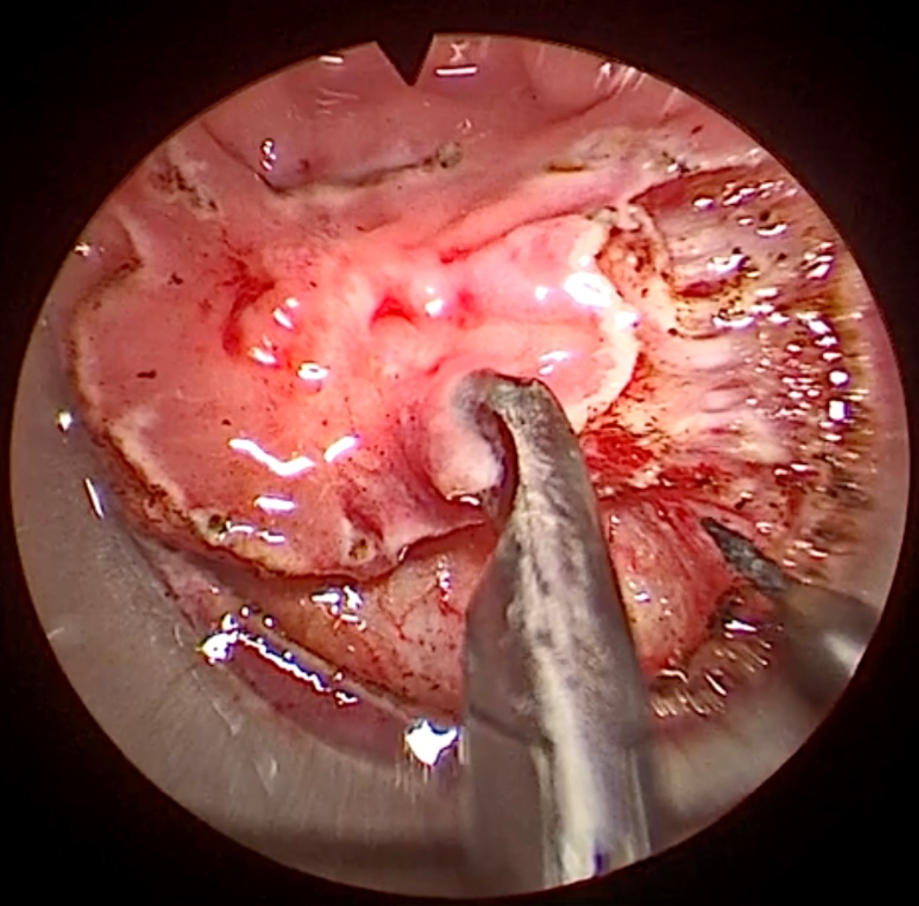}
            \caption*{D}
            \label{D}
        \end{subfigure} &
        \begin{subfigure}[b]{0.18\textwidth}
            \centering
            \includegraphics[height=3cm]{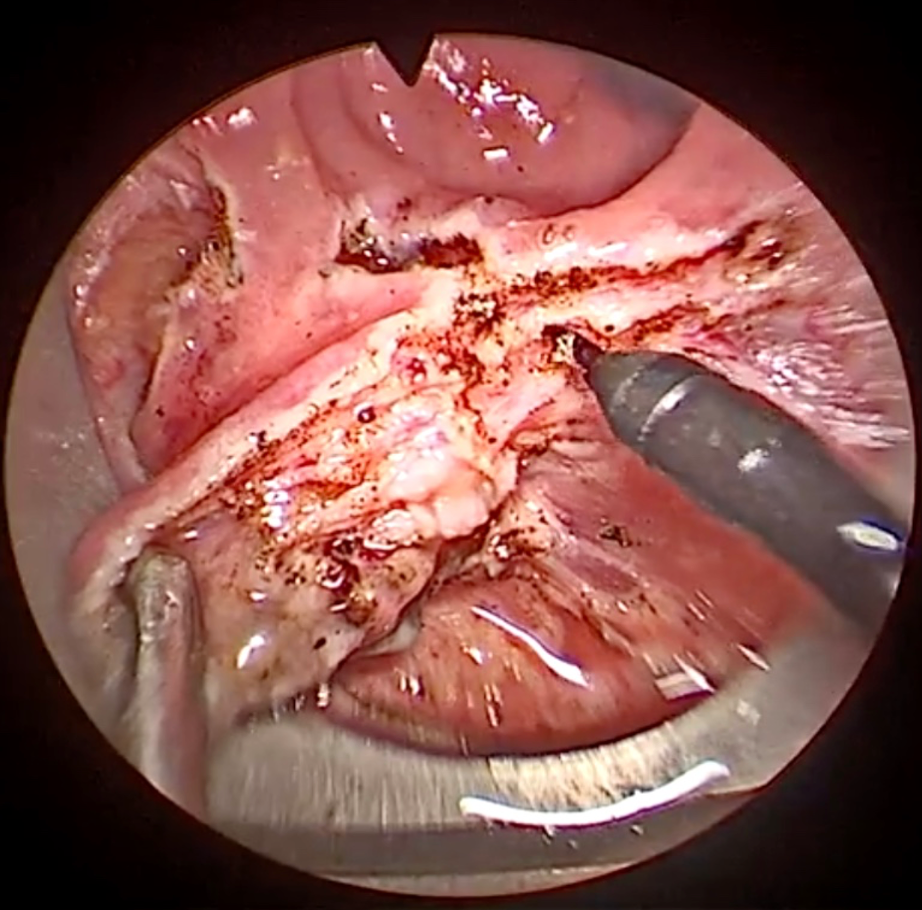}
            \caption*{E}
            \label{E}
        \end{subfigure} \\

        \begin{subfigure}[b]{0.18\textwidth}
            \centering
            \includegraphics[height=3cm]{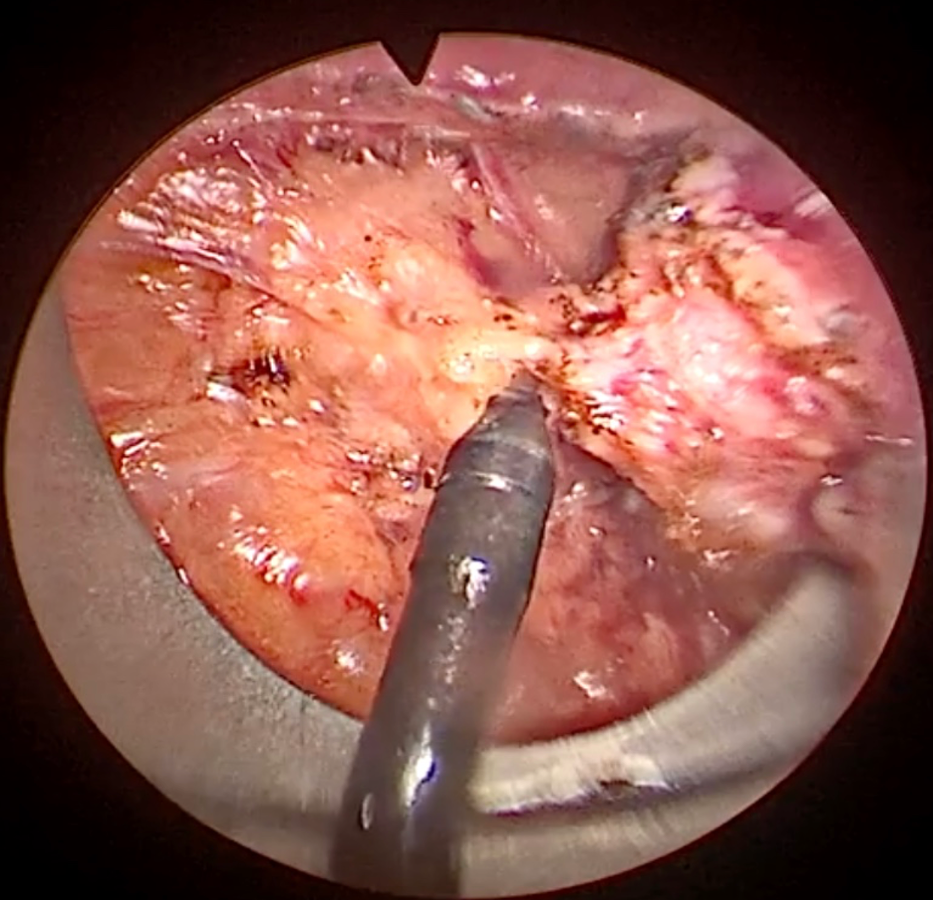}
            \caption*{F}
            \label{F}
        \end{subfigure} &
        \begin{subfigure}[b]{0.18\textwidth}
            \centering
            \includegraphics[height=3cm]{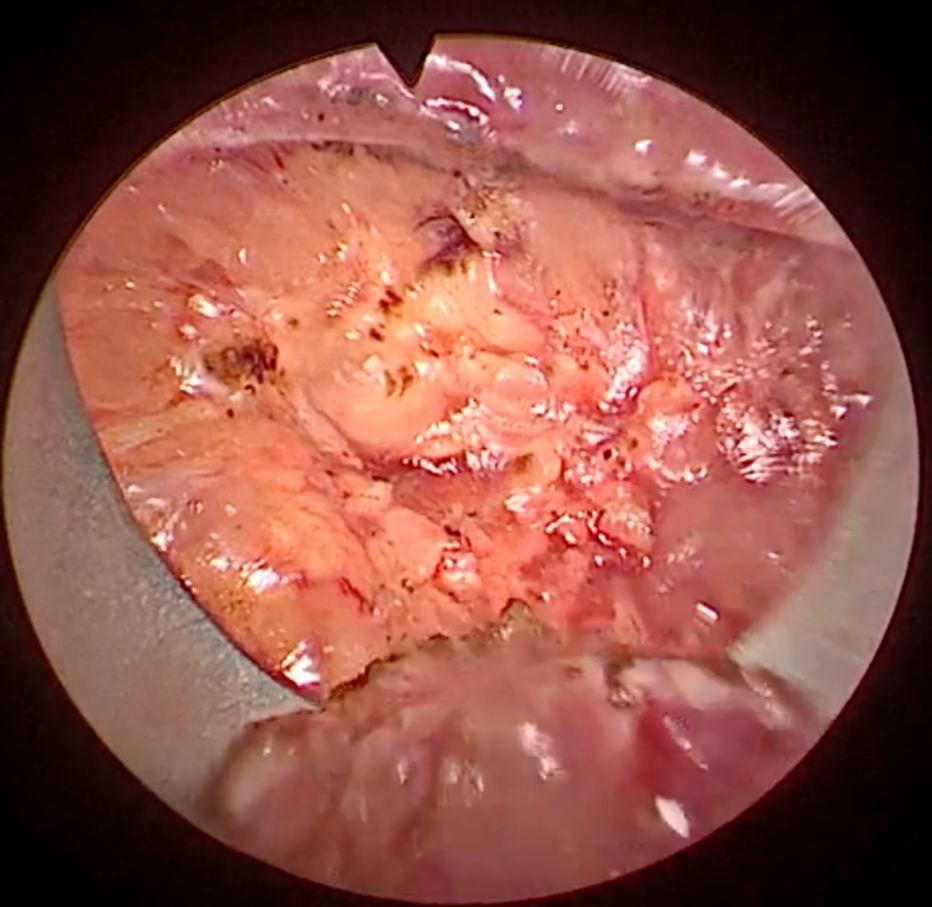}
            \caption*{G}
            \label{G}
        \end{subfigure} &
        \begin{subfigure}[b]{0.18\textwidth}
            \centering
            \includegraphics[height=3cm]{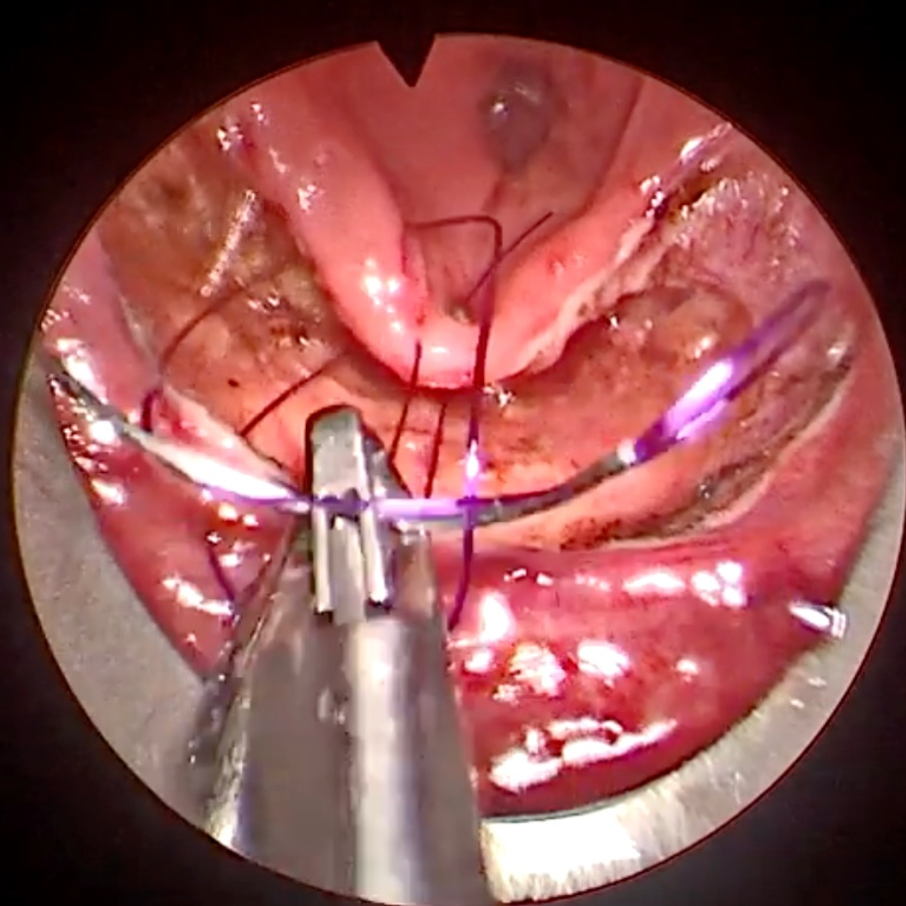}
            \caption*{H}
            \label{H}
        \end{subfigure} &
        \begin{subfigure}[b]{0.18\textwidth}
            \centering
            \includegraphics[height=3cm]{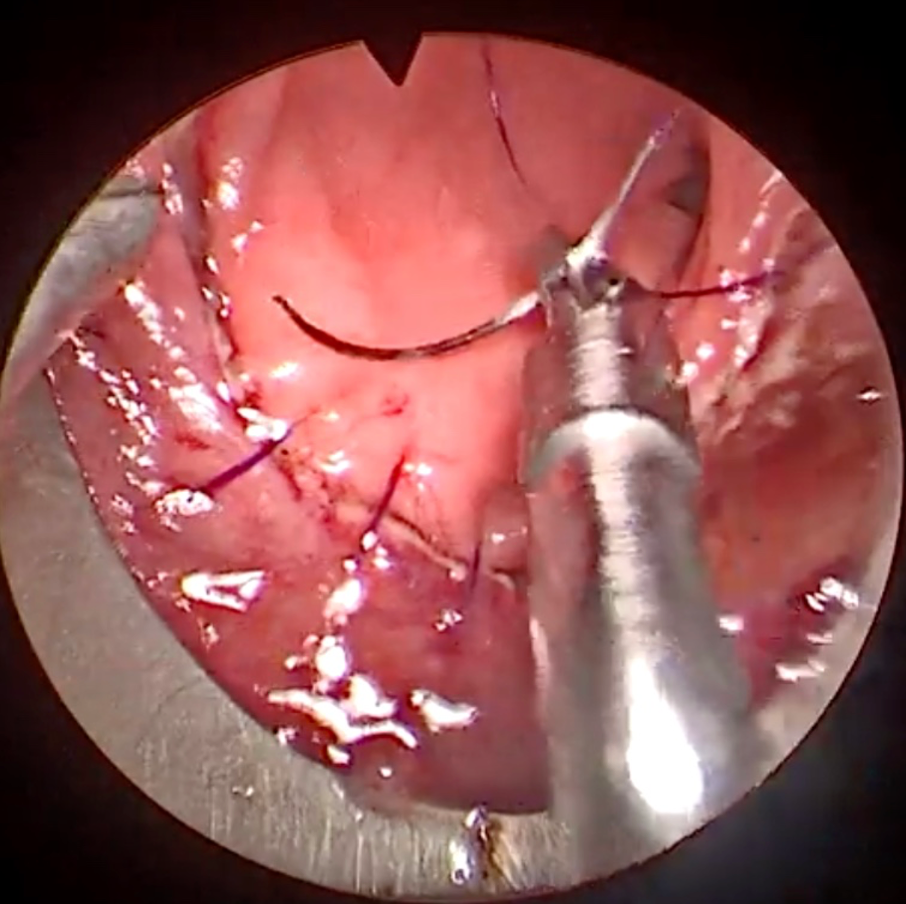}
            \caption*{I}
            \label{I}
        \end{subfigure} &
        \begin{subfigure}[b]{0.18\textwidth}
            \centering
            \includegraphics[height=3cm]{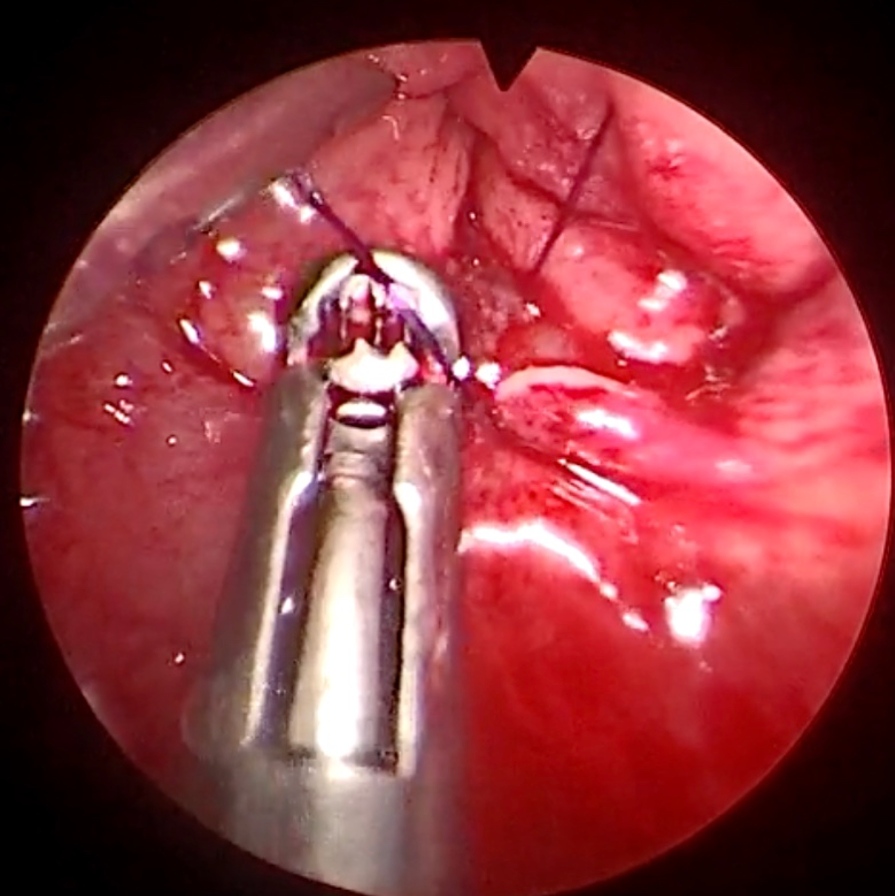}
            \caption*{J}
            \label{J}
        \end{subfigure}
    \end{tabular}
    \end{adjustbox}
    \caption{\textbf{TEMS surgical workflow:} \textit{A typical surgical flow from landmarking of the rectal polyp to dissection, lesion removal and closure of the rectal wall defect. The key milestones of a TEMS procedure are detailed in images A-J: [A] Baseline lesion in view after setup; [B] Application of landmark dots to outline the lesion; [C] Dissection of the wall through the mucosa and muscle; [D-E] Circumferential removal of the lesion; [F-G] Final removal and extraction of the specimen; [H-I] Closure of the rectal wall defect with a suture; and [J] Application of a metal clip to secure the suture and ensure complete closure.}}
    \label{fig:swfl}
\end{figure}


\vspace{2mm}
During dissection, multiple small events like surgical ``smoke'' fogging the lens, lens wash, tissue cauterisation, tissue retraction, fluid aspiration, and bleeding may occur. These are inter-related, for example tissue cauterisation results in surgical smoke that fogs up the operating scope and is normally managed by scope wash to clean the camera lens and aspiration of any fluid in the field of view. Bleeding is controlled with diathermy instruments and aspiration. Various instruments are used based on surgical needs. After dissection, the specimen is removed through the scope for histological analysis. In the closure phase, suturing to close the rectal wall defect. This involves handling the needle, driving it through the rectal wall, and pulling the suture to close the defect. Figure \ref{fig:swfl} illustrates the key steps of the surgical workflow for the entire TEMS procedure. 


\subsection*{Patient Cohort}

This study included fully de-identified videos from patients with a clinical diagnosis of rectal polyps or cancer. Pre-operatively, patients underwent standard clinical staging, including optical endoscopy, biopsy, endo-rectal ultrasound, magnetic resonance imaging, and computed tomography. These cases were discussed in a multidisciplinary meeting before elective surgery was offered. A team of four specialist colorectal surgeons, all Fellows of the Royal College of Surgeons (FRCS), performed TEMS using a Richard Wolf trans-anal operating platform.

\subsection*{Ethical Statement and Data Compliance}

The study is registered as a clinical audit with the University Hospitals Birmingham, conforming to local ethical standards, under the Clinical Audit Registration Management System (CARMS) number 20648. the surgical video analysis was performed by clinicians using scripts designed and shared by the computational data scientists.  Informed consent was obtained from all patients before recording fully de-identified surgical videos. Specifically, all patients signed institutional consent forms that gave permission to share a video on an open-access platform which may be seen by medical professionals or members of the general public. In accordance with NHS ethical standards and the UK General Data Protection Regulation (GDPR), the routinely collected ESV dataset underwent a full anonymisation procedure to ensure the removal of any identifiable information and protect patient privacy and confidentiality. Rigorous measures were implemented to review each video by clinicians to ensure that no patient identifiers were inadvertently captured or disclosed. Surgical scenes that extended beyond the abdominal cavity, capturing the surgical team or hospital/OR surroundings, were removed by the surgical team. These segments, typically occurring when the camera was temporarily extracted from the endo-luminal cavity for cleaning purposes, were replaced by blank frames while preserving patient privacy and maintaining the surgical procedure's overall chronological sequence and duration.

\subsection*{Data Capture \& Sharing}
We used the Operating Room (OR) visualisation system for data acquisition comprising a stereo endoscopic 50-degree scope and eyepiece attached to the Karl Storz Image 1 Hub HD (high-definition) camera system. Multi-part HD videos were recorded and archived using the Karl Storz AIDA™ system, which contains an intelligent export manager that automatically saves surgical video files during surgery. These files were stored on encrypted NHS hospital-based hard drives. All patient information was removed to ensure no metadata containing patient-related information was shared or made accessible to the clinical or data science teams.

\begin{figure}[!t]
    \centering
    \includegraphics[width=\textwidth, height=\textheight, keepaspectratio]{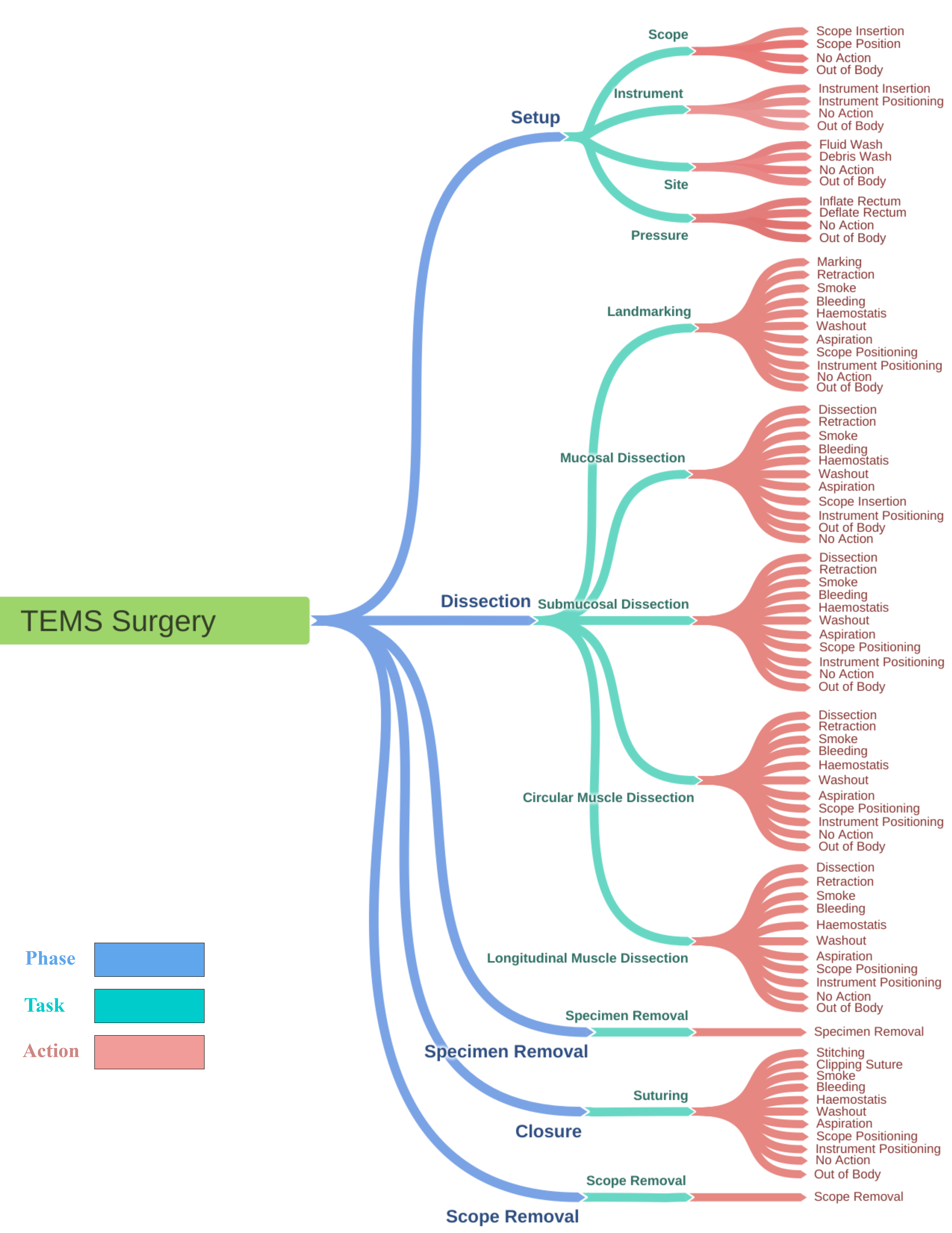}
    \caption{\textbf{Proposed Taxonomy of TEMS Surgical Workflow.} \textit{The TEMS operation can be split into three levels: [A] High level activity phase (such as Set-up, Dissection, Specimen Removal, Closure and Scope Removal), [B] Task based activities (such as scope insertion, instrument movement, site wash and pressure increase), [C] Small unit tasks (such as tissue marking, tissue retraction, smoke identification, bleeding identification and haemostasis).}}
    \label{fig:taxonomy}
\end{figure}

\subsection*{Co-Creation of Dense Taxonomy Labels for Timeline Segmentation}
In our effort to develop a comprehensive taxonomy for annotation, the project task group worked with specialist surgeons to define a representative surgical workflow. This collaborative co-creation process was essential to capturing the intricate details necessary to describe various downstream clinical tasks, in order to facilitate precise and detailed video labelling.

\vspace{2mm}
To achieve this, we structured the labels into \texttt{phase}, \texttt{task}, and \texttt{action} ``triplets''. This hierarchical framework allowed for a detailed end-to-end breakdown of the surgical procedure:
\begin{itemize}
    \item \textbf{Phase}: Represents ``high level'' activities encompassing a series of surgical tasks (for example setup of the TEMS scope or dissection phase).
    \item \textbf{Task}: Intermediate activities within a phase that include more specific actions (for example, in the dissection phase this may involve landmarking the dissection plane around the tumour or mucosal dissection).
    \item \textbf{Action}: The smallest unit activity within a task,  (for example, dissection, retraction, lens wash, identification of bleeding, haemostasis or aspiration of fluid).
\end{itemize}

\vspace{2mm}
For the TEMS procedure, we identified five key high-level phases: ``Setup'', ``Dissection'', ``Specimen Removal'', ``Closure of Defect'', and ``Scope Removal''. Each phase consists of various sub-tasks, which in turn are made up of individual specific actions. This ``triplet'' structure ensures that every aspect of the surgery is captured in detail, facilitating accurate and meaningful analysis. (See Figure \ref{fig:taxonomy} for detailed specification of our proposed TEMS surgical workflow taxonomy.)

\vspace{2mm}
This structured approach not only helps in the detailed documentation of the procedure but also enhances the ability to extract key events within an operation. This can be subsequently used to perform post-operative clinical assessments, either at an individual surgeon level or in comparison of techniques in a cohort of surgeons. This approach ensures that every critical operative action is accounted for and can be analysed in a constructive manner for measuring incremental performance improvements. This enables extremely ``dense'' data to be extracted from VBAs and analysed at scale. Large video libraries can be interrogated with the approaches defined above to understand changes in individual surgical performance over time as well as comparing surgeons against their peers.       

\begin{figure}[!t]
\begin{subfigure}{0.49\linewidth}
\includegraphics[width=\linewidth]{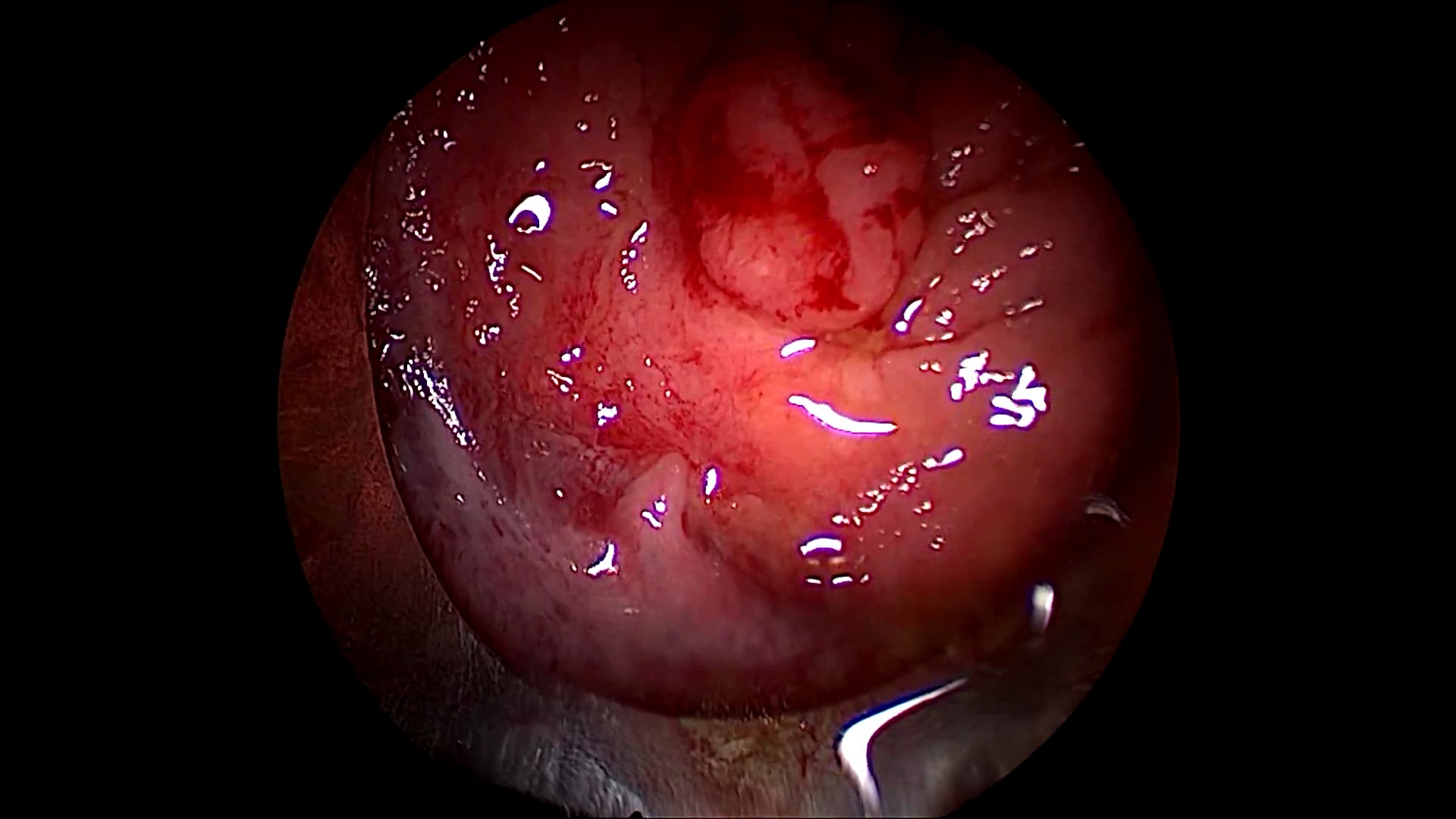}
\caption{Bitrate = 13 Mbps; Filesize = 1 GB.}
\label{fig:before_processing}
\end{subfigure}
\hfill
\begin{subfigure}{0.49\linewidth}
\includegraphics[width=\linewidth]{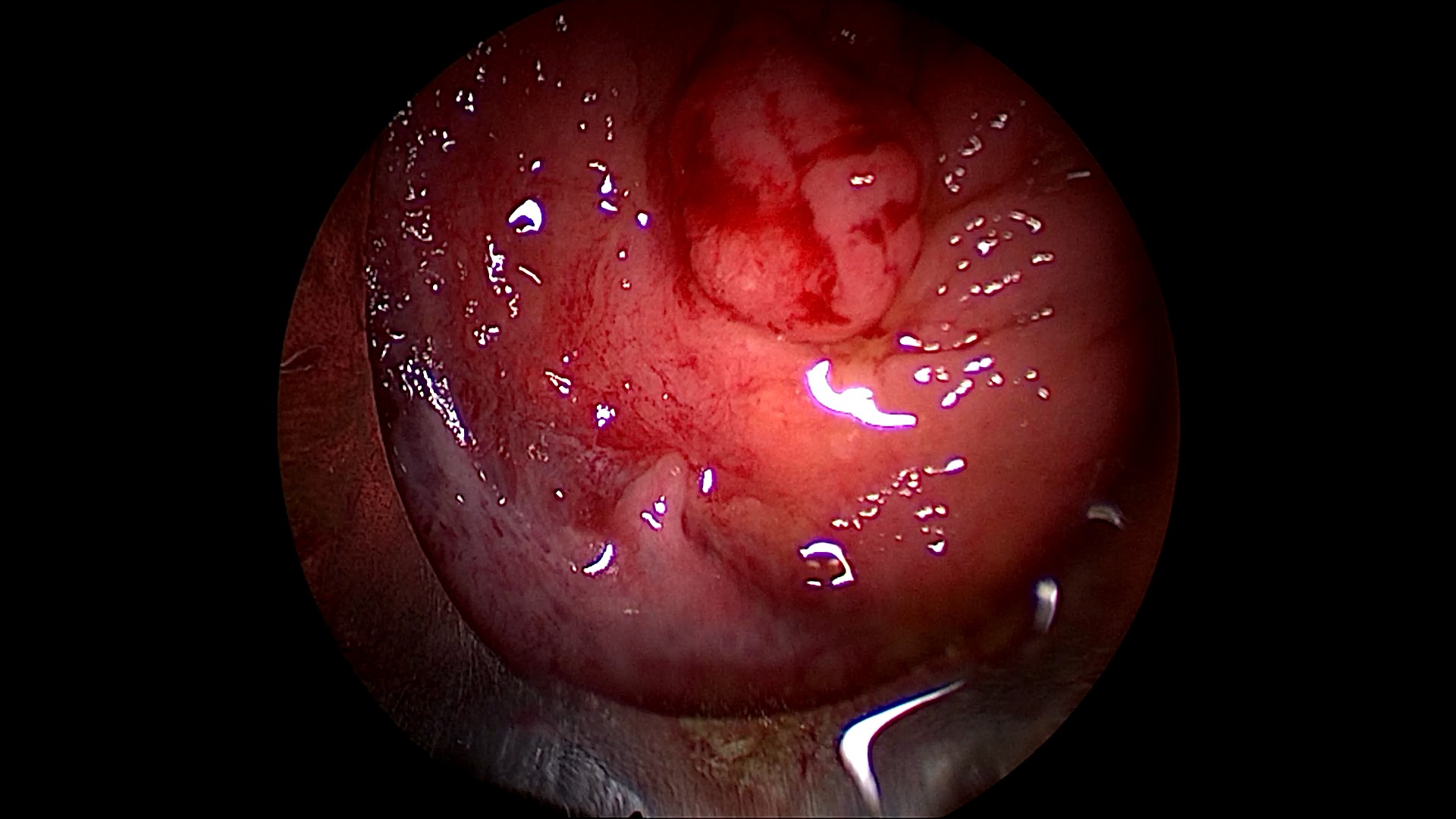}
\caption{Bitrate = 1 Mbps; Filesize = 100 MB.}
\label{fig:after_processing}
\end{subfigure}
\caption{\textbf{Side-by-side comparison of videos before and after pre-processing.} \textit{ Panels (a) and (b) show image quality before and after pre-processing, respectively. This shows that despite reduction in the size of the ESV file by a factor of 10 (from 1GB to 0.1GB), there was no loss in quality.}}
\end{figure}

\subsection*{Infrastructure Setup, Video Annotation, and Exporting Labels}
Our ESV dataset is extracted from multipart videos of TEMS surgeries, with uncompressed original videos of 10.34 gigabytes (GB) in size and covering procedures lasting up to 8 hours. We evaluated several labelling platforms and found LS (Label Studio version $1.12.1$) to be the most suitable for video annotation, despite some limitations. LS \cite {Gurevych2013-uo} is a secure web-based annotation tool supporting text, photos, videos, audio, and sequence data. While it offers extensive features, certain constraints affected its usability for our research. First, LS requires video alongside audio tracks for timeline segmentation functionality, but no audio was included in this dataset construction for obvious privacy reasons. Second, the default video upload limit for one file in LS is 250 megabytes, which can be extended but affects annotation interface performance adversely. Lastly, managing multi-part ESV files within a single project is challenging, as LS treats each surgery as a separate project, complicating data organisation.

\vspace{2mm}
To address these issues, we inserted blank audio tracks into the clips using \texttt{FFmpeg}\footnotemark{}\footnotetext{https://ffmpeg.org/}. We also reduced the file size by compressing the bitrate from 13 million (13 Mbps) to 100,000 (1 Mbps), facilitating smoother uploads and reducing server load without compromising visual quality. \Cref{fig:before_processing,fig:after_processing} shows a surgical scene before and after compression, demonstrating minimal quality loss.

\vspace{2mm}
After preparing the videos,  LS was installed on a server in a \texttt{Docker} container and made accessible for labelling by using a \texttt{ngrok} tunnelling platform. ESV files were uploaded into LS across several projects. For each project, the LS interface was customised for timeline annotation. The inclusion of empty audio tracks enabled the use of the LS timeline component for region-based labelling, significantly aiding efficient clinical annotation of large multi-part ESV files. Secure logins, allowed different surgeons to perform initial segmentation, which was then reviewed and finalised by a panel of clinical domain experts to ensure consistent labelling. Finally, we exported the labels from LS in JSON format, with each multi-part ESV file generating one JSON file for timeline segmentation.

\subsection*{Post-Processing of Annotations to Generate ML-ready Dataset}

We developed a systematic approach to transform the densely annotated multi-part ESV files into a machine learning-ready dataset. Instead of extracting all frames, we selected key frames using cosine distance similarity to capture significant scene changes in sequential order. This method reduced the number of extracted frames to an average of approximately $\sim550$ per video clip, compared to the total of around $\sim15K$ frames. Advanced \texttt{FFmpeg} features were utilised to uniquely name these keyframes by combining surgery and ESV file names with timestamps, ensuring chronological order. These keyframes are stored in the \texttt{frames} folder.

Next, we implemented a range-based query method to map each frame to its corresponding label from the Label Studio-exported JSON files. This mapping is saved in the \texttt{timeline\_labels.csv} file, which initially contains the columns \texttt{filename} and \texttt{timeline\_label\_raw}. To enhance the labelling data, we added additional columns such as \texttt{surgery\_name}, \texttt{video\_name}, and \texttt{timestamp}. We identified label overlaps at action transitions, where \texttt{timeline\_label\_raw} contained dual labels for boundary frames. To address this, we created a new \texttt{timeline\_label} column by selecting the trailing label and introduced a \texttt{transition} column to capture these timeline transitions. For multi-target modelling, we split \texttt{timeline\_label} into three columns: \texttt{timeline\_phase\_label}, \texttt{timeline\_task\_label}, and \texttt{timeline\_action\_label}, each representing different aspects of the TEMS taxonomy. Additionally, we added a \texttt{valid} column to indicate the suggested validation set, enabling the research community to benchmark algorithms using this dataset.

We also generated microclips for each frame, capturing the 30 seconds of video leading up to that frame. These microclips are stored as \texttt{.mp4} files in the \texttt{microclips} folder, using the same name as the corresponding frame.

The detailed steps of this post-processing approach are outlined in Algorithm~\ref{alg:post_processing}, which provides an overview of the key steps used to extract frames, map labels, and create microclips for timeline segmentation. The code is available in our GitHub repository \texttt{EVR}\footnotemark{}\footnotetext{\url{https://github.com/bilalcodehub/evr}}, where Python scripts such as \texttt{split.py}, \texttt{map.py}, and \texttt{microclip.py} can be used to carry out these tasks. Designed for flexibility, the \texttt{EVR} library can be applied to a wide range of surgical video datasets, ensuring systematic processing and transformation of raw video data and annotations into a structured format suitable for machine learning applications.

\begin{algorithm}[!t]
\caption{Post-Processing for Generating ESV Dataset}
\label{alg:post_processing} 
\KwData{ESV videos path, \texttt{offset} (microclip length)}
\KwResult{Extracted frames, timestamped labels, and corresponding microclips}

\Begin{
    \tcc{Step 1: Keyframe Extraction}
    \ForEach{video in dataset}{
        timestamps $\gets$ extract keyframe timestamps\;
        keyframes $\gets$ extract keyframes from video using timestamps\;
    }
    
    \tcc{Step 2: Load Annotations}
    annotation data $\gets$ load JSON annotations\;
    
    \tcc{Step 3: Label Mapping for Keyframes}
    \ForEach{keyframe in extracted keyframes}{
        video\_name, timestamp $\gets$ parse filename\;
        labels $\gets$ find labels from annotation data using video\_name and timestamp\;
        save keyframe filename, labels to results\;
    }
    
    \tcc{Step 4: Save Results}
    write results to \texttt{CSV}\;
    
    \tcc{Step 5: Create Microclips}
    \ForEach{entry in results}{
        video\_name, timestamp $\gets$ parse filename\;
        start\_time $\gets$ calculate start time (max(0, timestamp - \texttt{offset}))\;
        microclip\_file $\gets$ create microclip using \texttt{FFmpeg}\;
        save microclip to designated folder\;
    }
}
\end{algorithm}

\section*{Data Records}

We are releasing the \texttt{TEMSET-24K} dataset to the wider machine learning community to advance research and innovation in surgical data science. The dataset is hosted on \url{https://zenodo.org/records/14016844} under restricted access and requires an application to the UHB Research and Development PATHWAY team. The dataset is provided as a zipped \texttt{temset} folder, approximately 20 gigabytes in size, which includes several subfolders: \texttt{videos}, \texttt{frames}, and \texttt{microclips}.

\vspace{2mm}
The \texttt{videos} folder contains subfolders for the original video recordings of surgical procedures. Each subfolder is assigned a pseudonymised identifier, mapped to user-friendly names such as \texttt{TEMS-001}, \texttt{TEMS-002}, etc. Within each surgery folder (e.g., \texttt{TEMS-001}), there is a subfolder \texttt{originals} that holds the high-resolution, multi-part ESV files. These video clips are compressed using a bespoke data processing pipeline to approximately one-tenth of their original size, facilitating handling of large files. The compressed video clips are stored under the \texttt{videos} folder for ease of access and manipulation. The video clip names are retained post-compression to avoid ambiguity. Additionally, each surgery folder includes a JSON file (e.g., \texttt{TEMS-001.json}) exported from Label Studio, containing expert-annotated timeline segmentation labels using our proposed dense TEMS taxonomy. The JSON files are named according to the corresponding surgery folder, allowing uniform retrieval.

\vspace{2mm}
The \texttt{frames} folder contains the keyframes extracted by our algorithm to create a dataset suitable for machine learning tasks. These frames capture significant scene changes in the videos. The \texttt{microclips} folder contains 60-second video clips leading up to each keyframe, stored as \texttt{.mp4} files with names matching the corresponding frames.

\begin{figure}[!t]
    \centering
    \includegraphics[width=1\textwidth]{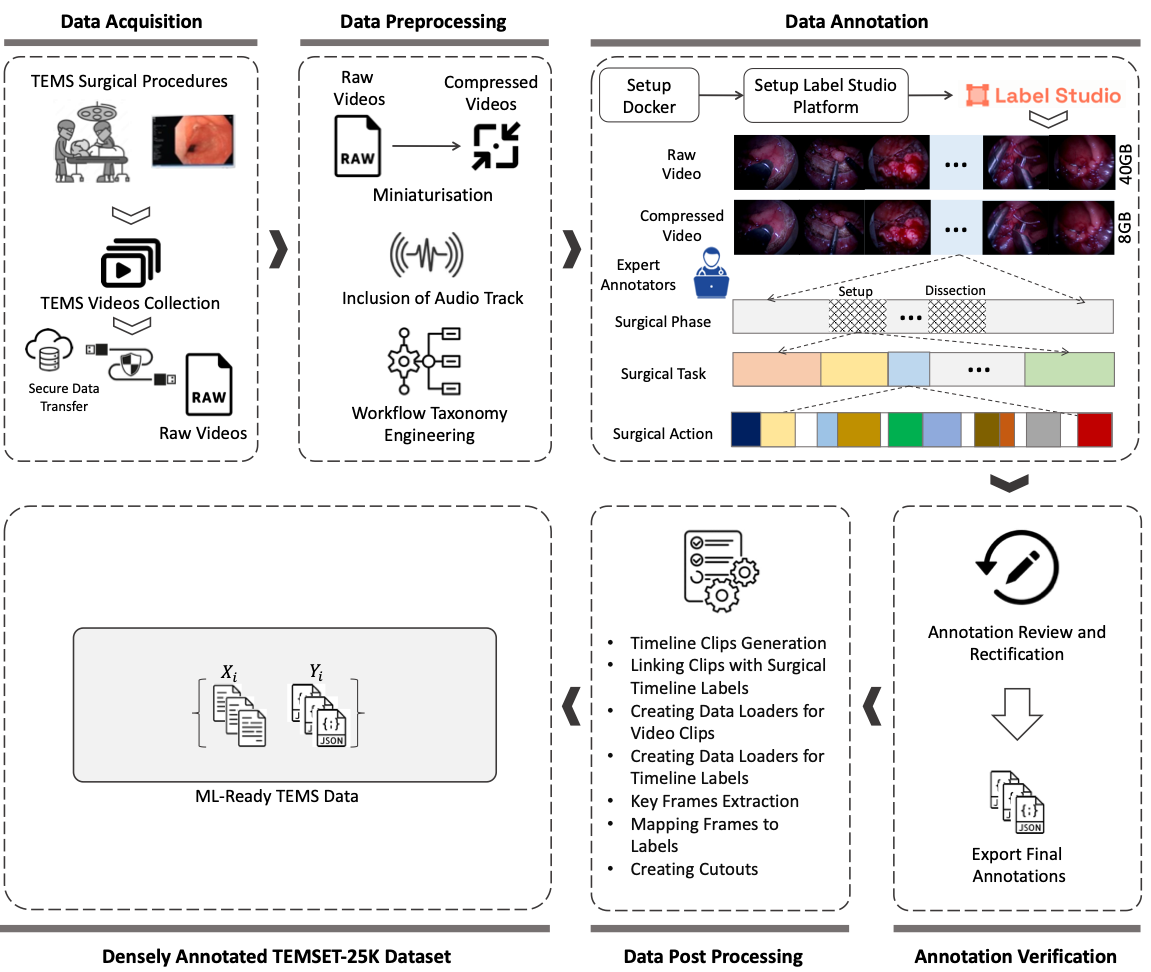}
    \caption{\textbf{Methodology adopted for annotating TEMS surgical videos for surgical timeline segmentation includes six major steps:} \textbf{(1)} \textit{TEMS surgical data acquisition \textbf{(2)} Data Preprocessing, \textbf{(3)} Data Annotation, \textbf{(4)} Annotation Verification, \textbf{(5)} Data Post Processing, and \textbf{(6)} Surgical data preparation for training timeline segmentation models. }}
    \label{fig:Timeline}
\end{figure}

\vspace{2mm}
The \texttt{timeline\_labels.csv} file, located in the \texttt{temset} folder, stores the labels for all microclips and frames. Each entry in the \texttt{filename} column comprises the surgical folder, video name, and timestamp, making each file uniquely identifiable and facilitating reproducibility by enabling tracking back to the original videos. The associated labels are presented both as dot-separated triplets and in individual columns for single and multi-target formulations. These labels are stored in the columns \texttt{timeline\_label}, \texttt{timeline\_phase\_label}, \texttt{timeline\_task\_label}, and \texttt{timeline\_action\_label}. Additionally, \texttt{timeline\_labels.csv} includes a \texttt{time\_to\_finish} column, which captures the remaining time of the surgery, calculated by subtracting the frame or microclip time from the total surgery duration. This facilitates research in models that can predict the remaining time for a procedure in surgical videos, offering numerous clinical applications.

\vspace{2mm}
Overall, \texttt{TEMSET-24K} includes $24,704$ expertly annotated microclips and frames, densely annotated by expert clinicians using timeline triplet labels. The entire data preparation, labelling, and construction pipeline is summarised in Fig.~\ref{fig:Timeline}.





\section*{Technical Validation}

\subsection*{Annotation Assessment}
To ensure the consistency of labelling in the dataset, we designed an annotation process involving a team of colorectal cancer surgery specialists, all accredited with fellowship status with the Royal College of Surgeons (RCS, UK). The process began with one surgeon annotating one full video in a shared setting to demonstrate the annotation procedure for the multi-part ESV files. Following this, another surgeon logged into the LS server using their credentials and navigated to the project they intended to annotate, accessing the individual video clips for annotation. The LS user interface provided a comma-separated list of phases, tasks, and actions for annotating the timeline of each video clip. Annotations were initially performed by one surgeon and subsequently validated by at least two other surgeons for cross-checking purposes. In cases of conflicting boundaries between the start and end of the labelling triplets, discussions were held to finalize the annotations that was agreed by all surgeons. We employed multifaceted strategies involving our proposed dense taxonomy, collaboratively annotating one full surgery in shared settings, and holding iterative discussions to resolve conflicts to achieve consistent annotations of the complex workflow scenes based on all surgeons' inputs. The final annotations consisted of labels made up of five phases, 12 tasks, and 21 actions as defined by the proposed taxonomy. These annotations were then programmatically exported from LS in JSON format, along with the corresponding ESV files.


%
\subsection*{Deep Learning Model Training}
\subsubsection*{Data Pre-Processing}
To improve the field of view, irrelevant areas were cropped from ESV images comprising black regions. The input image was first converted to grayscale, and a binary threshold was used to isolate the circular surgical region from the background. This step enhanced the visibility of the surgical scene. Subsequently, the largest contour was identified within the thresholded image and computed its minimum enclosing bounding box. A mask corresponding to this circular region was created and applied to the original image to extract the surgical area while ignoring the background. The bounding box of the surgical region was cropped and this cropped image was resized to its original size using bilinear interpolation. This method ensures that only the relevant surgical view is retained and standardised, facilitating improved visualisation and analysis of the surgical scene.

\begin{figure}[!t]
    \centering
    \includegraphics[width=1\textwidth]{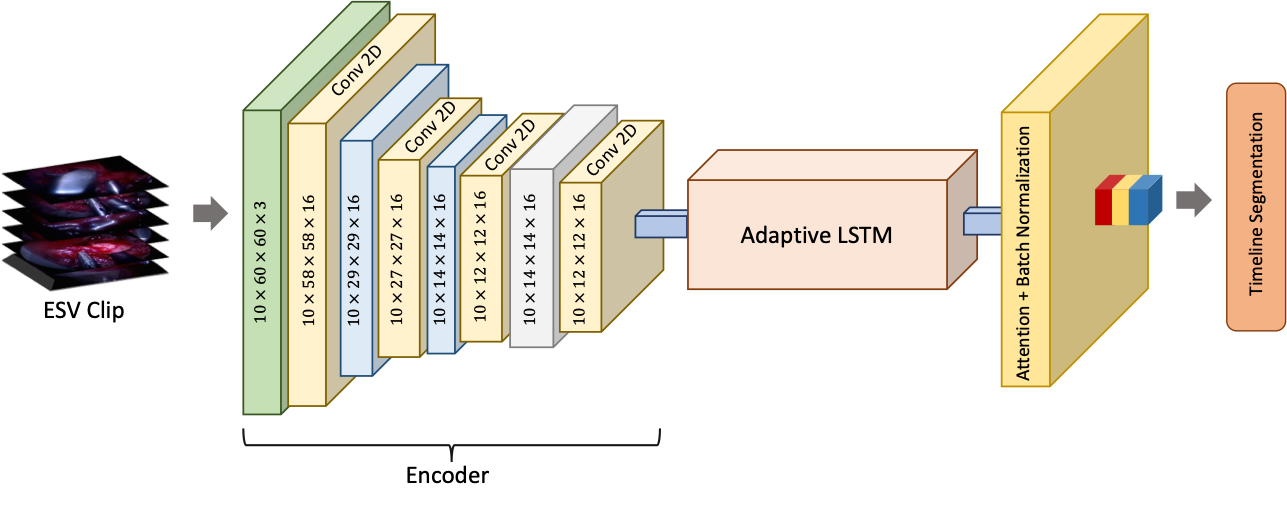}
    \caption{\textbf{Proposed SpatioTemporal Adaptive LSTM Network (STALNet) for Surgical Timeline Segmentation:} \textit{This network diagram shows the process by which ESV clips are analysed by encoders in order to apply reliable timeline segments.}}
    \label{fig:arch}
\end{figure}

\subsubsection*{Problem Formulation}
A key objective of this study was to learn an unknown function $\mathbf{F}$ that maps high-dimensional TEMS endoscopic surgical videos $\mathbf{X} \in \mathbb{R}^{T \times H \times W \times 3}$ to a multitarget label triplet $\mathbf{Y} \in \{\text{Phase}, \text{Task}, \text{Action}\}$, where, $T$, $H$, and $W$ denote the sequence length (no. of frames in the video), height, and width of the frames, respectively. To achieve this, this study proposes a Spatiotemporal Adaptive LSTM Network (STALNet) that learns the desired mapping. As shown in Fig. \ref{fig:arch}, STALNet integrates a TimeDistributed video encoder $\mathbf{E}^T$, followed by an adaptive long-short term memory network (LSTM) module having attention as the last layer $\mathbf{M}_\text{AA-LSTM}$ to capture spatial and temporal dependencies in the ESV data. Let $\phi$ be the feature extraction function using the backbone. The output of the encoder is given by:

\begin{equation}
    \mathbf{F} = \mathbf{E}^T(\phi(\mathbf{X})); \mathbf{X} \in \mathbb{R}^{B \times T \times C \times H \times W},
\end{equation}

where, $B$ is the batch size, $T$ is the sequence length, $C$ is the number of channels, and $H$ and $W$ are the height and width of the frames, respectively. We experimented with various encoders, including \texttt{ConvNeXt} (convnext\_small\_in22k) \cite{liu2022convnet}, \texttt{SWIN V2} (swinv2\_base\_window12\_192-22k)\cite{liu2022swin}, and \texttt{ViT} (vit\_small\_patch16\_224) \cite{steiner2021train,caron2021emerging}. These encoders were chosen for their proven ability to capture detailed spatial features across different scales, which is crucial for accurately interpreting surgical video frames. The extracted features are fed into an \texttt{Adaptive LSTM} module. This module consists of multiple LSTM layers, where the number of LSTMs depends on the input sequence length $T$. Each LSTM processes the sequence of features and produces hidden states. Let $\mathbf{h}_t$ represent the hidden state at time step $t$. The hidden states are computed as:
\[
\mathbf{H}_t = \mathbf{M}_\text{AA-LSTM}(\mathbf{F}_t, \mathbf{h}_{t-1}),
\]
where $\mathbf{H}_t \in \mathbb{R}^{B \times D}$. Multiple LSTM layers were applied to capture temporal dependencies across the sequence. Incorporating LSTMs into the proposed solution in an adaptive manner significantly improved the model's capacity for surgical scene understanding, as this approach leverages and preserves the temporal coherence in the videos, improving the stability and accuracy of the timeline predictions. The final hidden states from each LSTM layer are collected as $\mathbf{H} = [\mathbf{H}_1, \mathbf{H}_2, \ldots, \mathbf{H}_T] \in \mathbb{R}^{T \times B \times D}$ and their information across the sequence is aggregated using an attention mechanism. The attention weights are computed by applying a linear layer to the hidden states:
\[
\mathbf{A}_t = \text{softmax}(\mathbf{W}_a \mathbf{H}_t),
\]
where $\mathbf{W}_a \in \mathbb{R}^{D \times 1}$ is the attention weight matrix. The attention-weighted output is computed as a weighted sum of the hidden states:
\[
\mathbf{O} = \sum_{t=1}^{T} \mathbf{A}_t \mathbf{H}_t \in \mathbb{R}^{B \times D}.
\]

The final output is obtained by passing the attention-weighted output through a fully connected layer followed by batch normalisation:
\[
\mathbf{Y} = \text{BatchNorm}(\mathbf{W}_h \mathbf{O}),
\]
where $\mathbf{W}_h \in \mathbb{R}^{D \times (P+T+A)}$, with $P$, $T$, and $A$ representing the number of phases, tasks, and actions, respectively. A technique was employed here for mean ensembling to create more robust learners for each model, followed by heuristic-based prediction correction to address sporadic predictions. 

The model is trained using a custom loss function that combines the losses for phase, task, and action predictions. The total loss is given by:
\[
\mathcal{L} = \alpha \mathcal{L}_p + \beta \mathcal{L}_t + \gamma \mathcal{L}_a,
\]
where $\mathcal{L}_p$, $\mathcal{L}_t$, and $\mathcal{L}_a$ are the individual losses for phase, task, and action predictions, and $\alpha$, $\beta$, and $\gamma$ are their respective weights. Each of these losses is computed using the \texttt{CrossEntropyLossFlat} function applied to each of the output triplets.

\subsubsection*{DL Model Implementation}
The model described in this paper was implemented using the \texttt{fastai} \cite{howard2020fastai} library.  A server with 4 NVIDIA LS40 GPUs was used for training and validation. To enhance model convergence, the default \texttt{ReLU} activation function was replaced with the \texttt{Mish} activation function, which demonstrated superior performance in our experiments. Additionally, we substituted the default \texttt{Adam} optimiser with \texttt{ranger}, a combination of \texttt{RectifiedAdam} and the \texttt{Lookahead} optimisation technique, providing more stable and efficient training dynamics. To further optimise the training process, the \texttt{to\_fp16()} method was employed to reduce the precision of floating-point operations, thereby enabling half-precision training and improving computational efficiency. The \texttt{lr\_find} method was utilised to determine the optimal learning rate for the model, implementing a learning rate slicing technique. This approach assigned higher learning rates to the layers closer to the model head and lower learning rates to the initial layers, facilitating more effective training. For benchmarking, we initially evaluated several network architectures, including a basic image classifier, to establish a trivial baseline. This simple approach, however, produced significant sporadic predictions due to the absence of sequence modelling, highlighting the necessity for a more sophisticated model. 

\subsubsection*{Model Validation}
The model described in this paper was validated against the human annotator ground truth using the server with NVIDIA LS40 GPUs. We compared the proposed \texttt{STALNet} architecture with various encoder backbones, including \texttt{ConvNeXt}, \texttt{SWIN V2}, and \texttt{ViT}. The output results were analysed against the baseline to look at comparative performance metrics and how they captured the spatiotemporal dependencies that are crucial for the surgical timeline segmentation task. 

\subsubsection*{Statistical Analysis}
For our model evaluation, we utilised standard metrics including accuracy, F1 score, and ROC (Receiver Operating Characteristic) curves. To illustrate model variability, standard deviation is reported for accuracy and F1 scores. The following equations define these metrics: 

\begin{equation}
\resizebox{\textwidth}{!}{$\text{Accuracy} = \frac{TP + TN}{TP + TN + FP + FN} \times 100\%,
\text{Precision} = \frac{TP}{TP + FP},
\text{Recall} = \frac{TP}{TP + FN},
\text{F1 Score} = 2 \cdot \frac{\text{Precision} \cdot \text{Recall}}{\text{Precision} + \text{Recall}}.$}
\end{equation}
We computed these statistics at two levels: 1) Overall Model Performance: We reported the overall accuracy and F1 score on the entire validation set. 2) Class-Specific Performance: These metrics were computed for each taxonomy triplet class (phase, task, and action) to identify which classes the model struggles with the most. Additionally, ROC curves were used to visually investigate model performance. True positives (TP), false positives (FP), true negatives (TN), and false negatives (FN) were derived from the predictions, which were then used to compute precision and recall, leading to the construction of ROC curves plotted using Scikit-learn. To enhance our analysis, we implemented custom visualisations showing video clips, target labels, and model predictions. We employed color coding (red for incorrect and green for correct predictions) for easy interpretation. All data and model results were visualised and analysed using Matplotlib, NumPy, and Scikit-learn.







\begin{table}[!t]
\caption{Comparison of Surgical Timeline Segmentation Models}
\centering
\label{Tab-comparison}
\begin{adjustbox}{max width=0.7\textwidth}
\begin{tabular}{l|l|rr|rr|rr}
\hline
\rowcolor[HTML]{CBCEFB} 
\cellcolor[HTML]{CBCEFB}                                & \cellcolor[HTML]{CBCEFB}                                            & \multicolumn{2}{c|}{\cellcolor[HTML]{CBCEFB}\textbf{ConvNeXt}}                                                                  & \multicolumn{2}{c|}{\cellcolor[HTML]{CBCEFB}\textbf{ViT}}                                                                       & \multicolumn{2}{c}{\cellcolor[HTML]{CBCEFB}\textbf{SWIN V2}}                                                                   \\ \cline{3-8} 
\rowcolor[HTML]{CBCEFB} 
\multirow{-2}{*}{\cellcolor[HTML]{CBCEFB}\textbf{Sr\#}} & \multirow{-2}{*}{\cellcolor[HTML]{CBCEFB}\textbf{Model Descripton}} & \multicolumn{1}{c|}{\cellcolor[HTML]{CBCEFB}\textbf{Accuracy}} & \multicolumn{1}{c|}{\cellcolor[HTML]{CBCEFB}\textbf{F1 Score}} & \multicolumn{1}{c|}{\cellcolor[HTML]{CBCEFB}\textbf{Accuracy}} & \multicolumn{1}{c|}{\cellcolor[HTML]{CBCEFB}\textbf{F1 Score}} & \multicolumn{1}{c|}{\cellcolor[HTML]{CBCEFB}\textbf{Accuracy}} & \multicolumn{1}{c}{\cellcolor[HTML]{CBCEFB}\textbf{F1 Score}} \\ \hline
1                                                       & Baseline Vision Classifier                                          & \multicolumn{1}{r|}{80.36\%}                                   & 72.99\%                                                        & \multicolumn{1}{r|}{75.23\%}                                   & 60.87\%                                                        & \multicolumn{1}{r|}{78.74\%}                                   & 66.70\%                                                       \\ \hline
2                                                       & STALNet (Ours)                                                      & \multicolumn{1}{r|}{\textbf{91.69\%}}                          & 82.78\%                                                        & \multicolumn{1}{r|}{83.02\%}                                   & 68.29\%                                                        & \multicolumn{1}{r|}{91.42\%}                                   & \textbf{86.02\%}                                              \\ \hline
\end{tabular}
\end{adjustbox}
\end{table}

\subsubsection*{Model Performance Evaluation}
Table \ref{Tab-comparison} presents the accuracy and F1 scores for each model across the three encoder architectures. The baseline image classification learner, which predicts timeline labels based solely on individual images, achieved an F1 score of 72.99\% with the ConvNeXt encoder, 66.7\% with the SWIN V2 encoder, and 60.87\% with the ViT encoder. These results indicate the fundamental capability of deep learning models for surgical timeline segmentation but also highlight the limitations of relying solely on spatial information. In contrast, our proposed \texttt{STALNet} demonstrated significant performance improvements over the baseline model. On average, \texttt{STALNet} achieved an F1 score of 82.78\% and an accuracy of 91.69\%, reflecting an average performance gain of 9.79\% in F1 score and 11.38\% in accuracy compared to the baseline model. These improvements underscore the importance of incorporating spatiotemporal information for surgical timeline segmentation. Furthermore, the performance varied between different model encoders used in the time-distributed layer for feature extraction. Among the evaluated encoders, the ConvNeXt encoder achieved the highest accuracy with 91.69\%, slightly better than the SWIN V2 encoder at 91.41\%. However, the highest performing F1 score, which is a significant metric for evaluating timeline segmentation, was achieved by the SWIN V2 encoder at 86.02\%, which is approximately 3.24\% higher than the ConvNeXt encoder's F1 score of 82.78\%. This demonstrates that while ConvNeXt offers marginally better accuracy, SWIN V2 excels in terms of F1 score, highlighting its superior performance in capturing relevant features for timeline segmentation. Despite the higher F1 score of SWIN V2, it required substantial computation during both training and deployment phases. On the other hand, ConvNeXt not only delivered competitive performance but also offered a more computationally efficient solution, making it a practical choice for real-world applications. Overall, the \texttt{STALNet} model, particularly with the ConvNeXt encoder, demonstrated superior performance in segmenting surgical timelines. This highlights the efficacy of integrating spatiotemporal features and selecting robust encoder architectures to balance performance and computational efficiency.

\begin{figure}[h]
    \centering
    \begin{subfigure}[b]{0.32\textwidth}
        \centering
        \includegraphics[width=\textwidth, height=4cm]{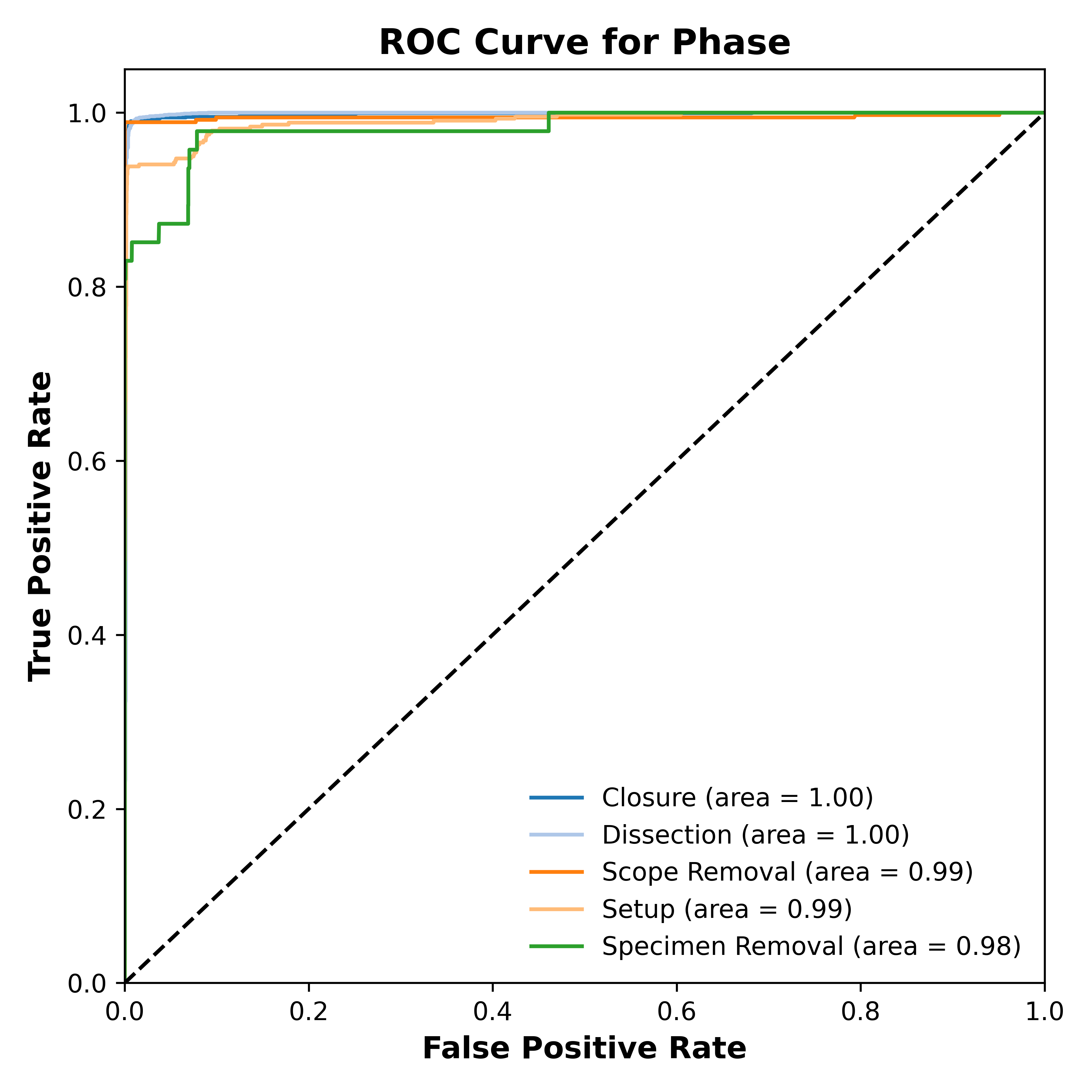}
        \caption{ConvNeXt}
        \label{ConvNextPhase}
    \end{subfigure}
    \hfill
    \begin{subfigure}[b]{0.32\textwidth}
        \centering
        \includegraphics[width=\textwidth, height=4cm]{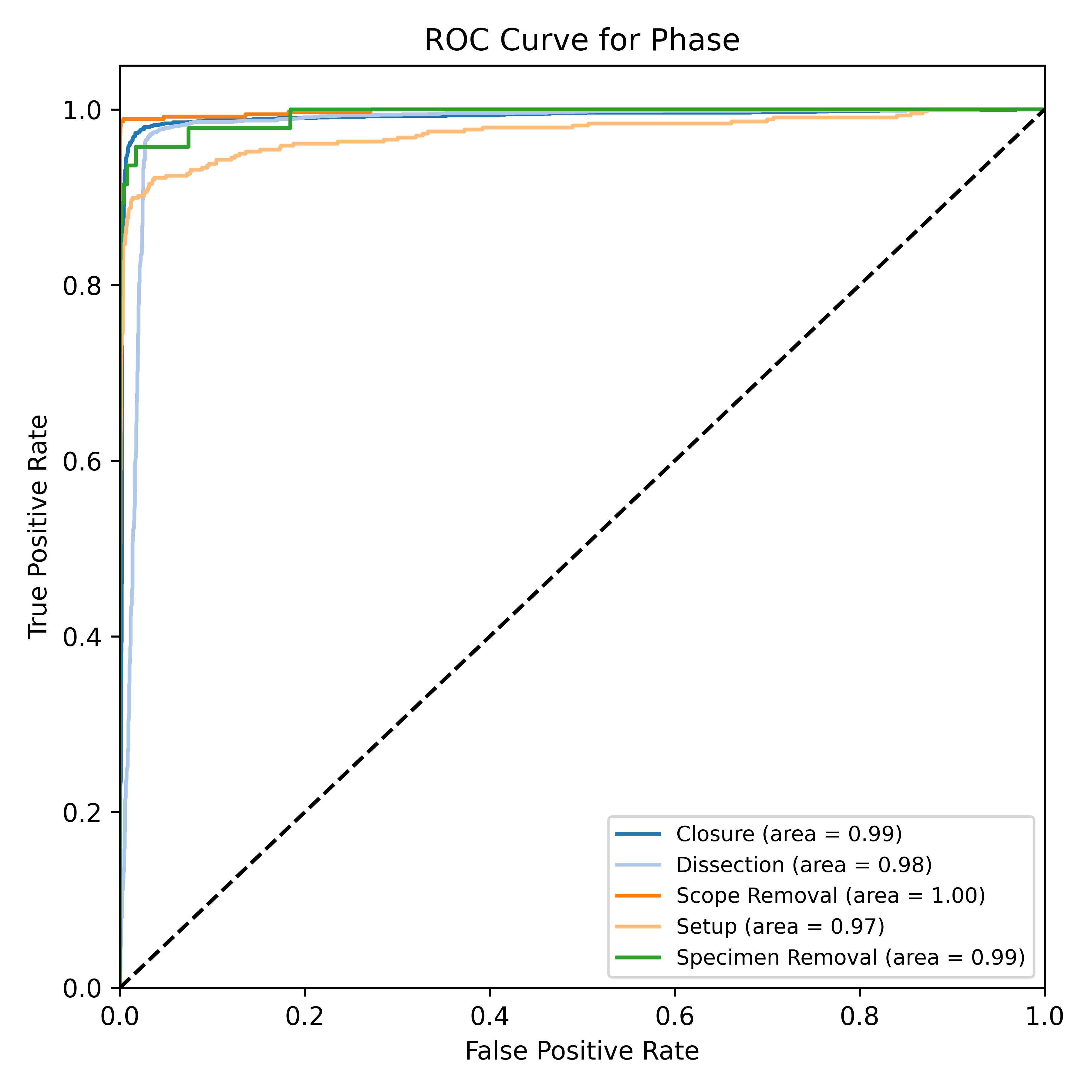}
        \caption{ViT}
        \label{ViTPhase}
    \end{subfigure}
    \hfill
    \begin{subfigure}[b]{0.32\textwidth}
        \centering
        \includegraphics[width=\textwidth, height=4cm]{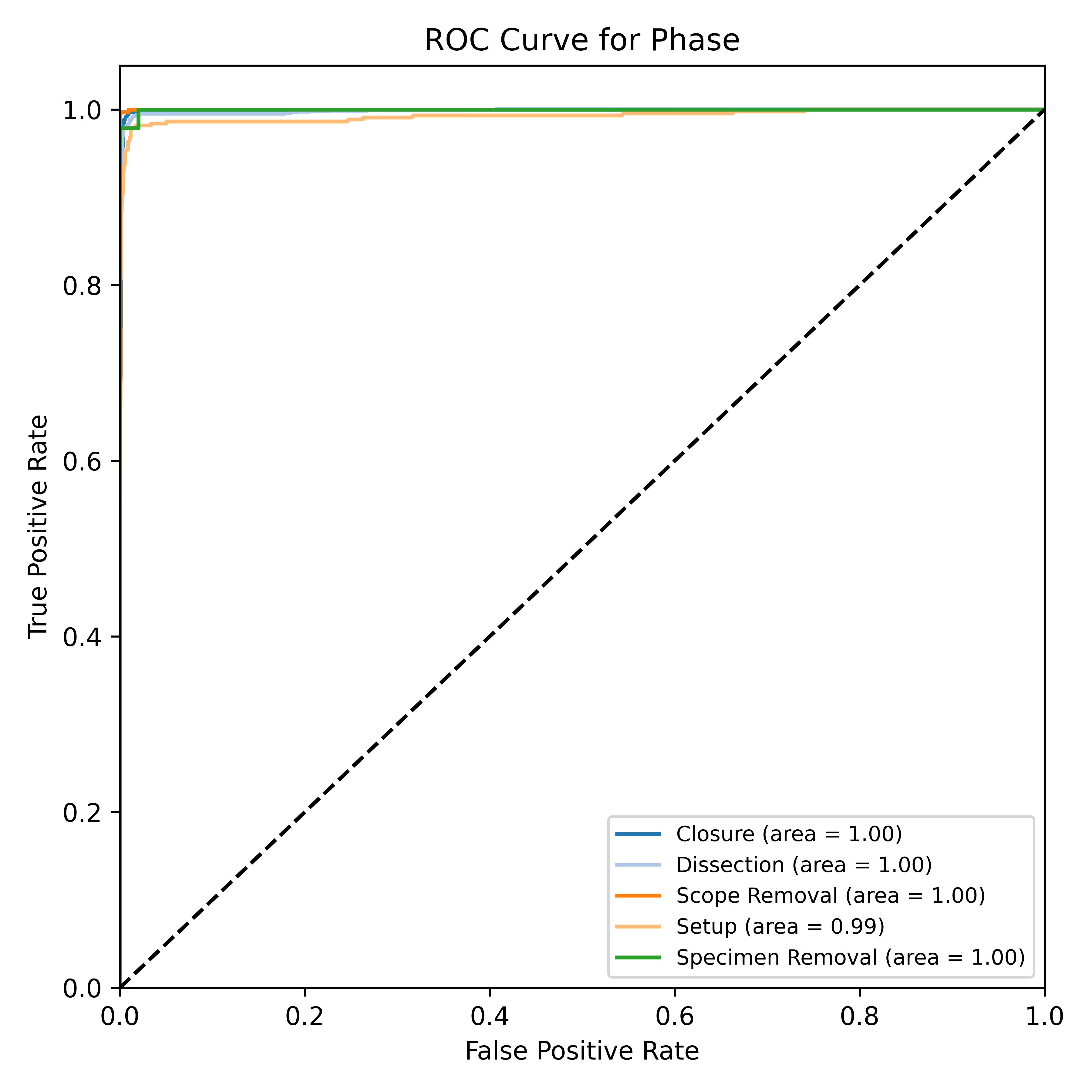}
        \caption{SWIN V2}
        \label{SWIN V2Phase}
    \end{subfigure}
 \hfill
    \begin{subfigure}[b]{0.32\textwidth}
        \centering
        \includegraphics[width=\textwidth, height=4cm]{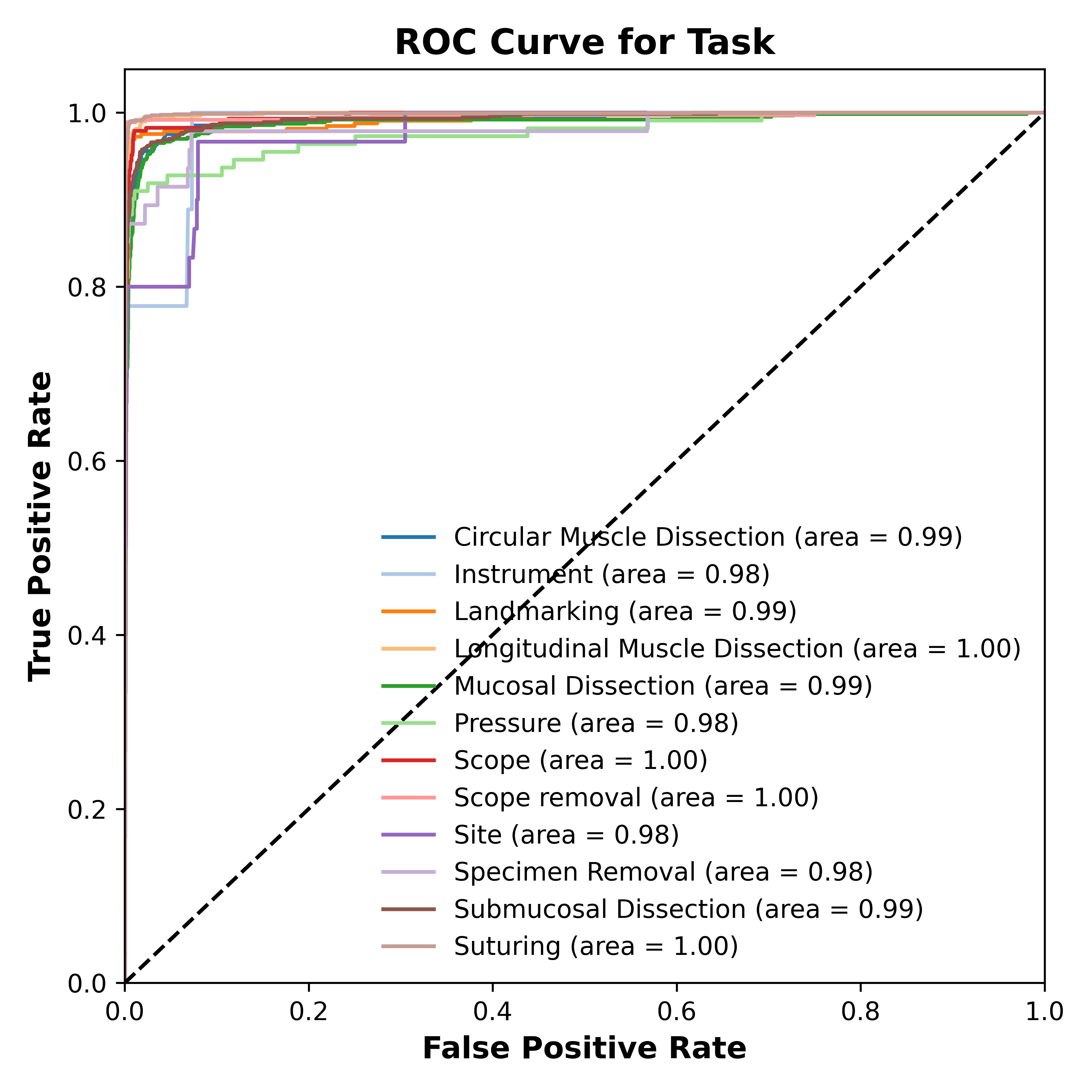}
        \caption{ConvNeXt}
        \label{ConvNextTask}
    \end{subfigure}
    \hfill
    \begin{subfigure}[b]{0.32\textwidth}
        \centering
        \includegraphics[width=\textwidth, height=4cm]{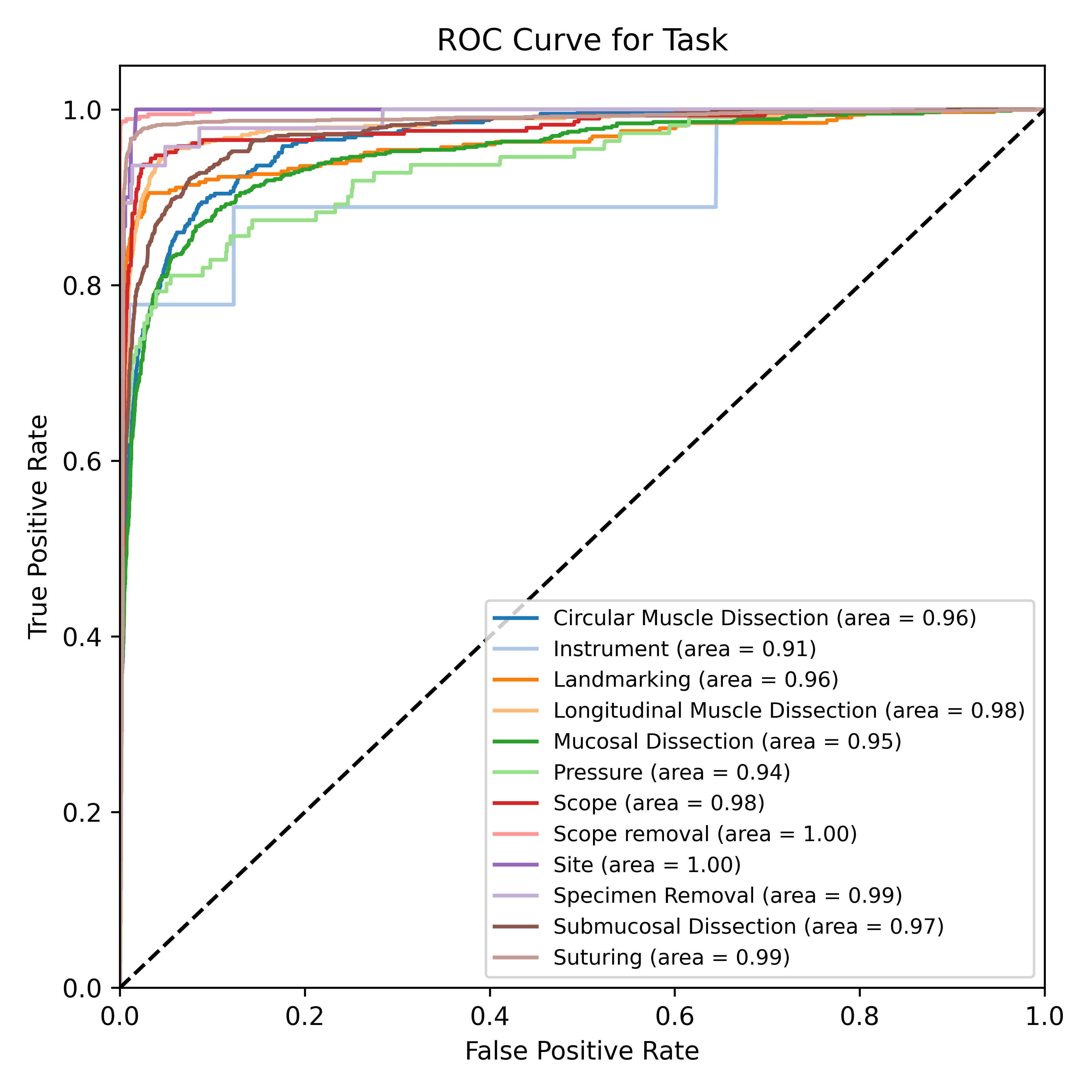}
        \caption{ViT}
        \label{ViTTask}
    \end{subfigure}
    \hfill
    \begin{subfigure}[b]{0.32\textwidth}
        \centering
        \includegraphics[width=\textwidth, height=4cm]{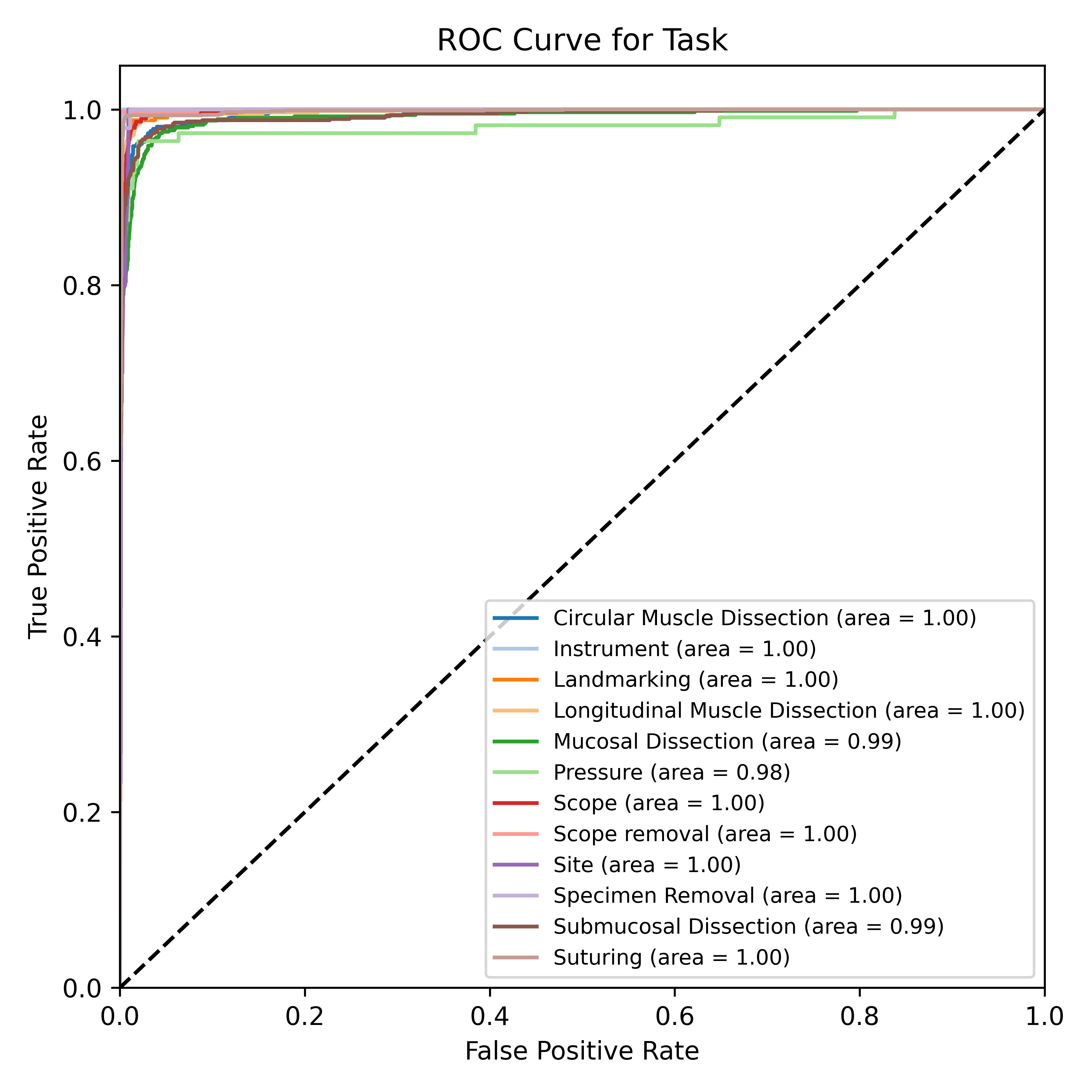}
        \caption{SWIN V2}
        \label{SWIN V2}
    \end{subfigure}
    \begin{subfigure}[b]{0.32\textwidth}
        \centering
        \includegraphics[width=\textwidth, height=4cm]{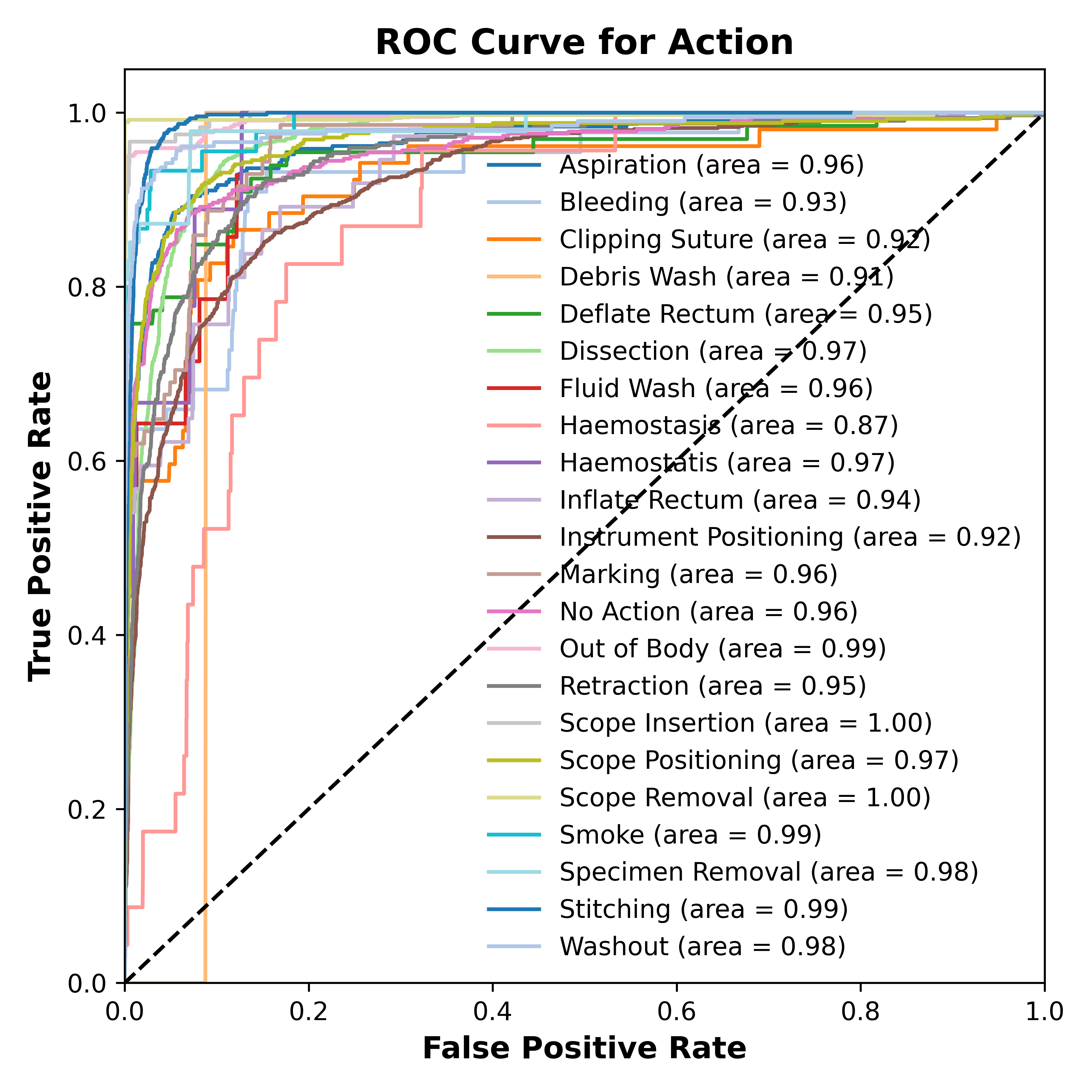}
        \caption{ConvNeXt}
        \label{ConvNextAction}
    \end{subfigure}
    \hfill
    \begin{subfigure}[b]{0.32\textwidth}
        \centering
        \includegraphics[width=\textwidth, height=4cm]{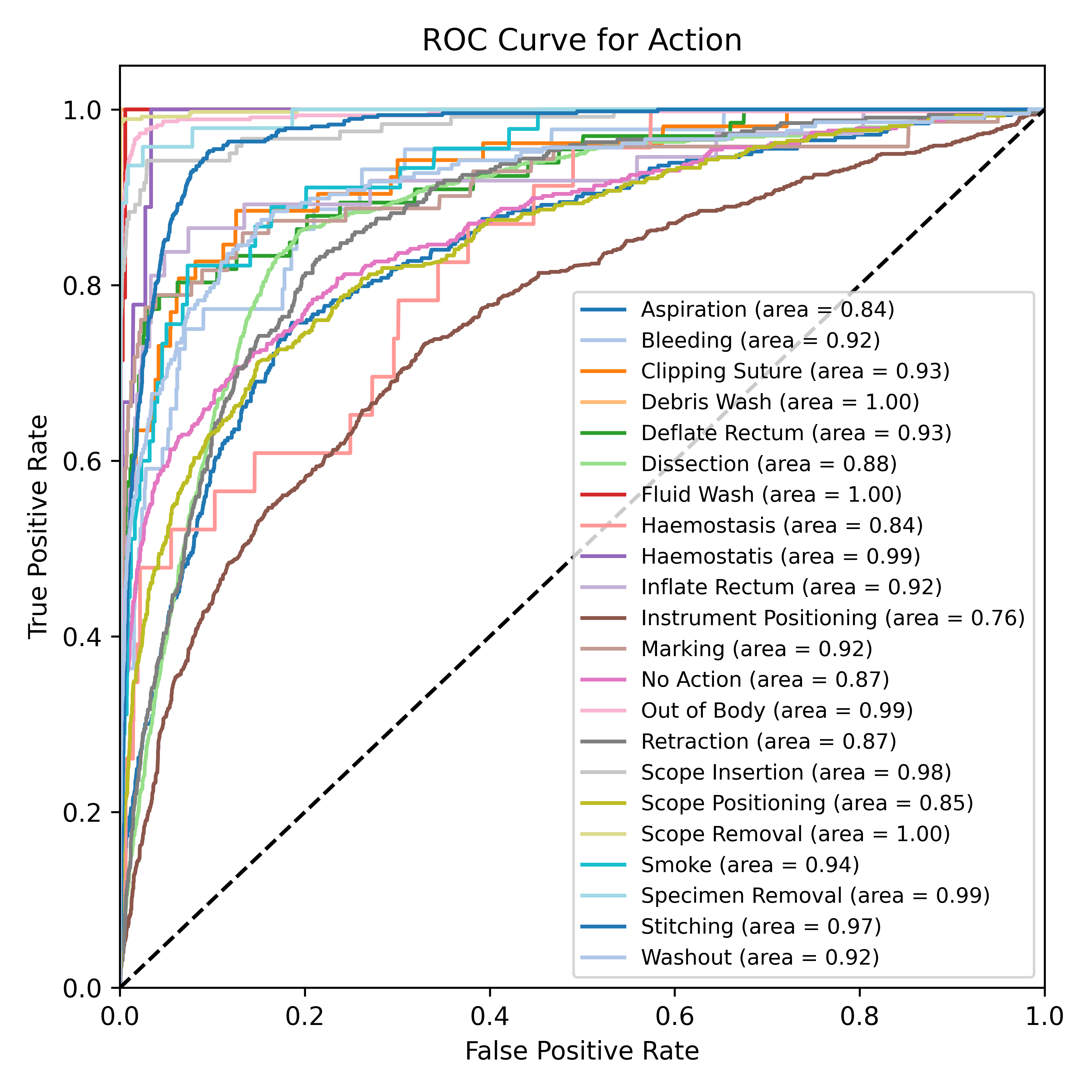}
        \caption{ViT}
        \label{ViT}
    \end{subfigure}
    \hfill
    \begin{subfigure}[b]{0.32\textwidth}
        \centering
        \includegraphics[width=\textwidth, height=4cm]{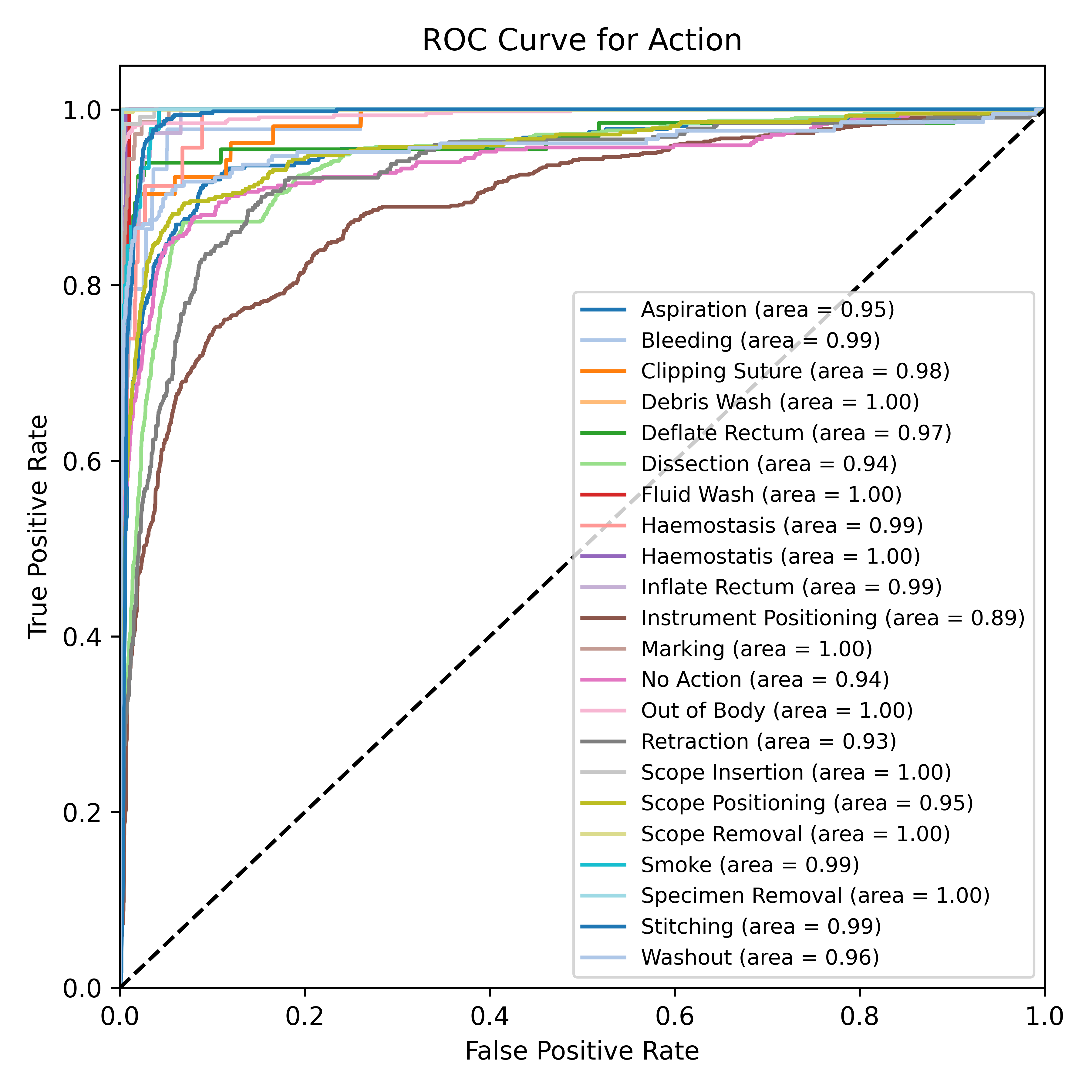}
        \caption{SWIN V2}
        \label{SWIN V2Action}
    \end{subfigure}
    \caption{\textbf{STALNet Performance Review using ROC Curves for Taxonomy Triplets}: \textit{The top row of ROC curves shows the performance of ConvNeXt, ViT and SWIN V2 encoders on labelling high level TEMS surgical "Phases". The next two rows show the performance of STALNet encoders on labelling TEMS surgical "Tasks" (intermediate level) and "Actions" (the fine level)}.}
    \label{fig:roc}
\end{figure}

\vspace{2mm}
The \texttt{STALNet} model was also evaluated for its performance on each of the taxonomy triplets (phase, task, action) as shown in \cref{Tab-phs}, \cref{Tab-task}, and  \cref{Tab-act}, respectively. The evaluation of phase segmentation reveals that the model performs exceptionally well across all phases, with only minor fluctuations in performance using different encoders. The ROC curves show its efficacy across these triplet behaviours (see Figures \ref{fig:roc}). For example, the “Dissection” phase achieved an F1 score of 99.0\% with no variance and an accuracy of 99.0\% with a variance of 11.0\% with the SWIN V2 encoder. Similarly, the “Setup” phase showed high performance with an F1 score of 98.0\% and an accuracy of 99.0\%, both exhibiting low variances (1\% and 9\%, respectively with ConvNeXt and SWIN V2 encoders). Even the “Closure” phase, despite being one of the more challenging phases due to its fewer instances, maintained an F1 score and accuracy of 100\% for both with variances of 0\% and 5\%, respectively with the SWIN V2 encoder. These results indicate that the model effectively captures and segments the different phases consistently across three distinct encoders. In task segmentation, the model showed strong and consistent performance across most tasks. For instance, tasks such as “Longitudinal Muscle Dissection” and “Suturing” achieved high F1 scores of 99\% for each, with accuracies of 100\% and 99\%, and low variances (1\% and 0\%, and 7\% and 8\%, respectively) with the ConvNeXt encoder. This consistency reflects the model's robust ability to segment tasks accurately. Conversely, the “Site” task, which had a significantly lower F1 score of 67\% with high variance 33\% with the ConvNeXt encoder. This indicates that the model struggles more with tasks that are less frequently represented in the dataset. For action segmentation, the model demonstrated high performance on frequently occurring actions such as “Scope Insertion” and “Stitching” achieving F1 scores of 99\% and 95\%, and accuracies of 100\% and 98\%, respectively with the ConvNeXt encoder. The variances for “Scope Insertion” were 1\% for the F1 score and 3\% for accuracy, while “Stitching” had variances of 4\% and 15\%, indicating stable and reliable performance. However, actions like “Debris Wash” and “Haemostatis,” which had lower F1 scores of 50\% for each, also exhibited higher variances 50\% for each of the above actions with the ConvNeXt encoder . These findings suggest that the model's performance is consistent for well-represented actions, but struggles with less frequent actions. 

\begin{table}[!t]
\centering
\caption{Performance of the STALNet model on Surgical Phases across different encoders}
\label{Tab-phs}
\renewcommand{\arraystretch}{1.2}
\begin{adjustbox}{max width=0.7\textwidth}
\begin{tabular}{p{4.8cm}|c|c|c|c|c|c}
\hline
\rowcolor[HTML]{CBCEFB} 
\cellcolor[HTML]{CBCEFB}                                & \multicolumn{2}{c|}{\cellcolor[HTML]{CBCEFB}\textbf{ConvNeXt}}                                                                  & \multicolumn{2}{c|}{\cellcolor[HTML]{CBCEFB}\textbf{ViT}}                                                                       & \multicolumn{2}{c}{\cellcolor[HTML]{CBCEFB}\textbf{SWIN V2}}                                                                   \\ \cline{2-7} 
\rowcolor[HTML]{CBCEFB} 
\multirow{-2}{*}{\cellcolor[HTML]{CBCEFB}\textbf{Phase Name}} & \multicolumn{1}{c|}{\cellcolor[HTML]{CBCEFB}\textbf{Accuracy}} & \multicolumn{1}{c|}{\cellcolor[HTML]{CBCEFB}\textbf{F1 Score}} & \multicolumn{1}{c|}{\cellcolor[HTML]{CBCEFB}\textbf{Accuracy}} & \multicolumn{1}{c|}{\cellcolor[HTML]{CBCEFB}\textbf{F1 Score}} & \multicolumn{1}{c|}{\cellcolor[HTML]{CBCEFB}\textbf{Accuracy}} & \multicolumn{1}{c}{\cellcolor[HTML]{CBCEFB}\textbf{F1 Score}} \\ \hline
[01] Setup                                                   & \multicolumn{1}{r|}{0.99 ±  0.09}                              & 0.97 ±  0.02                                                   & \multicolumn{1}{r|}{0.98 ±  0.13}                              & 0.94 ±  0.05                                                   & \multicolumn{1}{r|}{0.99 ±  0.10}                              & 0.97 ±  0.03                                                  \\ \hline
[02] Dissection                                              & \multicolumn{1}{r|}{0.99 ±  0.10}                              & 0.99 ±  0.00                                                   & \multicolumn{1}{r|}{0.97 ±  0.17}                              & 0.97 ±  0.00                                                   & \multicolumn{1}{r|}{0.99 ±  0.11}                              & 0.99 ±  0.00                                                  \\ \hline

[03] Specimen Removal                                        & \multicolumn{1}{r|}{1.00 ±  0.03}                              & 0.97 ±  0.03                                                   & \multicolumn{1}{r|}{1.00 ±  0.04}                              & 0.95 ±  0.05                                                   & \multicolumn{1}{r|}{1.00 ±  0.02}                              & 0.99 ±  0.01                                                  \\ \hline
[04] Closure                                                 & \multicolumn{1}{r|}{0.99 ±  0.09}                              & 0.99 ±  0.00                                                   & \multicolumn{1}{r|}{0.98 ±  0.14}                              & 0.98 ±  0.01                                                   & \multicolumn{1}{r|}{0.99 ±  0.08}                              & 0.99 ±  0.00                                                  \\ \hline
[05] Scope Removal                                           & \multicolumn{1}{r|}{1.00 ±  0.03}                              & 1.00 ±  0.00                                                   & \multicolumn{1}{r|}{1.00 ±  0.05}                              & 0.99 ±  0.01                                                   & \multicolumn{1}{r|}{1.00 ±  0.03}                              & 1.00 ±  0.00                                                  \\ \hline

\end{tabular}
\end{adjustbox}
\end{table}

\begin{table}[!t]
\centering
\caption{Performance of the STALNet model on Surgical Tasks across different encoders}
\label{Tab-task}
\renewcommand{\arraystretch}{1.2}
\begin{adjustbox}{max width=0.7\textwidth}
\begin{tabular}{p{5.1cm}|c|c|c|c|c|c}
\hline
\rowcolor[HTML]{CBCEFB} 
\cellcolor[HTML]{CBCEFB} &
  \multicolumn{2}{c|}{\cellcolor[HTML]{CBCEFB}\textbf{ConvNeXt}} &
  \multicolumn{2}{c|}{\cellcolor[HTML]{CBCEFB}\textbf{ViT}} &
  \multicolumn{2}{c}{\cellcolor[HTML]{CBCEFB}\textbf{SWIN V2}} \\ \cline{2-7} 
\rowcolor[HTML]{CBCEFB} 
\multirow{-2}{*}{\cellcolor[HTML]{CBCEFB}\textbf{Task Name}} &
  \multicolumn{1}{c|}{\cellcolor[HTML]{CBCEFB}\textbf{Accuracy}} &
  \multicolumn{1}{c|}{\cellcolor[HTML]{CBCEFB}\textbf{F1 Score}} &
  \multicolumn{1}{c|}{\cellcolor[HTML]{CBCEFB}\textbf{Accuracy}} &
  \multicolumn{1}{c|}{\cellcolor[HTML]{CBCEFB}\textbf{F1 Score}} &
  \multicolumn{1}{c|}{\cellcolor[HTML]{CBCEFB}\textbf{Accuracy}} &
  \multicolumn{1}{c}{\cellcolor[HTML]{CBCEFB}\textbf{F1 Score}} \\ \hline
  
[01] Scope Setup &
  \multicolumn{1}{r|}{0.99 ± 0.10} &
  0.96 ± 0.04 &
  \multicolumn{1}{r|}{0.99 ± 0.11} &
  0.94 ± 0.05 &
  \multicolumn{1}{r|}{0.99 ± 0.09} &
  0.96 ± 0.04 \\ \hline
  [02] Instrument Setup &
  \multicolumn{1}{r|}{1.00 ± 0.02} &
  0.94 ± 0.06 &
  \multicolumn{1}{r|}{1.00 ± 0.03} &
  0.81 ± 0.19 &
  \multicolumn{1}{r|}{1.00 ± 0.02} &
  0.92 ± 0.08 \\ \hline
  [03] Site Setup &
  \multicolumn{1}{r|}{1.00 ± 0.06} &
  0.83 ± 0.17 &
  \multicolumn{1}{r|}{1.00 ± 0.07} &
  0.84 ± 0.16 &
  \multicolumn{1}{r|}{1.00 ± 0.07} &
  0.82 ± 0.18 \\ \hline
  [04] Pressure Setup &
  \multicolumn{1}{r|}{0.99 ± 0.07} &
  0.93 ± 0.07 &
  \multicolumn{1}{r|}{0.99 ± 0.11} &
  0.85 ± 0.15 &
  \multicolumn{1}{r|}{0.99 ± 0.08} &
  0.93 ± 0.07 \\ \hline
[05] Landmarking &
  \multicolumn{1}{r|}{0.99 ± 0.07} &
  0.98 ± 0.02 &
  \multicolumn{1}{r|}{0.98 ± 0.13} &
  0.93 ± 0.06 &
  \multicolumn{1}{r|}{0.99 ± 0.09} &
  0.97 ± 0.03 \\ \hline
[06] Mucosal Dissection &
  \multicolumn{1}{r|}{0.98 ± 0.14} &
  0.95 ± 0.04 &
  \multicolumn{1}{r|}{0.94 ± 0.23} &
  0.87 ± 0.10 &
  \multicolumn{1}{r|}{0.98 ± 0.15} &
  0.95 ± 0.04 \\ \hline
[07] Submucosal Dissection &
  \multicolumn{1}{r|}{0.98 ± 0.14} &
  0.96 ± 0.03 &
  \multicolumn{1}{r|}{0.96 ± 0.20} &
  0.91 ± 0.06 &
  \multicolumn{1}{r|}{0.98 ± 0.14} &
  0.96 ± 0.03 \\ \hline
[08] Circular Muscle Dissection &
  \multicolumn{1}{r|}{0.99 ± 0.11} &
  0.96 ± 0.04 &
  \multicolumn{1}{r|}{0.96 ± 0.19} &
  0.86 ± 0.12 &
  \multicolumn{1}{r|}{0.98 ± 0.12} &
  0.95 ± 0.04 \\ \hline
[09] Longitudinal Muscle Dissection &
  \multicolumn{1}{r|}{0.99 ± 0.11} &
  0.97 ± 0.02 &
  \multicolumn{1}{r|}{0.97 ± 0.17} &
  0.93 ± 0.06 &
  \multicolumn{1}{r|}{0.99 ± 0.11} &
  0.97 ± 0.02 \\ \hline
[10] Specimen Removal &
  \multicolumn{1}{r|}{1.00 ± 0.03} &
  0.98 ± 0.02 &
  \multicolumn{1}{r|}{1.00 ± 0.04} &
  0.96 ± 0.04 &
  \multicolumn{1}{r|}{1.00 ± 0.02} &
  0.99 ± 0.01 \\ \hline
[11] Suturing &
  \multicolumn{1}{r|}{0.99 ± 0.09} &
  0.99 ± 0.00 &
  \multicolumn{1}{r|}{0.98 ± 0.14} &
  0.98 ± 0.01 &
  \multicolumn{1}{r|}{0.99 ± 0.09} &
  0.99 ± 0.00 \\ \hline
[12] Scope removal &
  \multicolumn{1}{r|}{1.00 ± 0.03} &
  1.00 ± 0.00 &
  \multicolumn{1}{r|}{1.00 ± 0.05} &
  0.99 ± 0.01 &
  \multicolumn{1}{r|}{1.00 ± 0.03} &
  1.00 ± 0.00 \\ \hline

\end{tabular}
\end{adjustbox}
\end{table}

\begin{table}[!t]
\centering
\caption{Performance of the STALNet model on Surgical Actions across different encoders}
\label{Tab-act}
\renewcommand{\arraystretch}{1.2}
\begin{adjustbox}{max width=0.7\textwidth}
\begin{tabular}{p{5.1cm}|c|c|c|c|c|c}
\hline
\rowcolor[HTML]{CBCEFB} 
\cellcolor[HTML]{CBCEFB} &
  \multicolumn{2}{c|}{\cellcolor[HTML]{CBCEFB}\textbf{ConvNeXt}} &
  \multicolumn{2}{c|}{\cellcolor[HTML]{CBCEFB}\textbf{ViT}} &
  \multicolumn{2}{c}{\cellcolor[HTML]{CBCEFB}\textbf{SWIN V2}} \\ \cline{2-7} 
\rowcolor[HTML]{CBCEFB} 
\multirow{-2}{*}{\cellcolor[HTML]{CBCEFB}\textbf{Action Name}} &
  \multicolumn{1}{c|}{\cellcolor[HTML]{CBCEFB}\textbf{Accuracy}} &
  \multicolumn{1}{c|}{\cellcolor[HTML]{CBCEFB}\textbf{F1 Score}} &
  \multicolumn{1}{c|}{\cellcolor[HTML]{CBCEFB}\textbf{Accuracy}} &
  \multicolumn{1}{c|}{\cellcolor[HTML]{CBCEFB}\textbf{F1 Score}} &
  \multicolumn{1}{c|}{\cellcolor[HTML]{CBCEFB}\textbf{Accuracy}} &
  \multicolumn{1}{c}{\cellcolor[HTML]{CBCEFB}\textbf{F1 Score}} \\ \hline
[01] Aspiration             & \multicolumn{1}{r|}{0.97 ± 0.18} & 0.85 ± 0.13 & \multicolumn{1}{r|}{0.93 ± 0.26} & 0.69 ± 0.27 & \multicolumn{1}{r|}{0.97 ± 0.18} & 0.85 ± 0.13 \\ \hline
[02] Bleeding               & \multicolumn{1}{r|}{0.99 ± 0.07} & 0.82 ± 0.17 & \multicolumn{1}{r|}{0.99 ± 0.10} & 0.67 ± 0.33 & \multicolumn{1}{r|}{0.99 ± 0.08} & 0.80 ± 0.19 \\ \hline
[03] Clipping Suture        & \multicolumn{1}{r|}{1.00 ± 0.07} & 0.88 ± 0.12 & \multicolumn{1}{r|}{0.99 ± 0.10} & 0.71 ± 0.28 & \multicolumn{1}{r|}{1.00 ± 0.06} & 0.91 ± 0.09 \\ \hline
[04] Debris Wash            & \multicolumn{1}{r|}{1.00 ± 0.01} & 0.50 ± 0.50 & \multicolumn{1}{r|}{1.00 ± 0.01} & 0.50 ± 0.50 & \multicolumn{1}{r|}{1.00 ± 0.01} & 0.50 ± 0.50 \\ \hline
[05] Deflate Rectum         & \multicolumn{1}{r|}{0.99 ± 0.07} & 0.89 ± 0.10 & \multicolumn{1}{r|}{0.99 ± 0.10} & 0.77 ± 0.23 & \multicolumn{1}{r|}{0.99 ± 0.08} & 0.89 ± 0.11 \\ \hline
[06] Dissection             & \multicolumn{1}{r|}{0.93 ± 0.26} & 0.87 ± 0.08 & \multicolumn{1}{r|}{0.87 ± 0.34} & 0.76 ± 0.16 & \multicolumn{1}{r|}{0.93 ± 0.26} & 0.86 ± 0.10 \\ \hline
[07] Fluid Wash             & \multicolumn{1}{r|}{1.00 ± 0.05} & 0.76 ± 0.24 & \multicolumn{1}{r|}{1.00 ± 0.06} & 0.78 ± 0.22 & \multicolumn{1}{r|}{1.00 ± 0.06} & 0.73 ± 0.27 \\ \hline
[08] Haemostatis            & \multicolumn{1}{r|}{1.00 ± 0.03} & 0.85 ± 0.15 & \multicolumn{1}{r|}{1.00 ± 0.03} & 0.79 ± 0.21 & \multicolumn{1}{r|}{1.00 ± 0.03} & 0.85 ± 0.15 \\ \hline
[09] Inflate Rectum         & \multicolumn{1}{r|}{1.00 ± 0.07} & 0.83 ± 0.17 & \multicolumn{1}{r|}{0.99 ± 0.08} & 0.70 ± 0.29 & \multicolumn{1}{r|}{1.00 ± 0.06} & 0.87 ± 0.13 \\ \hline
[10] Instrument Positioning & \multicolumn{1}{r|}{0.91 ± 0.29} & 0.82 ± 0.12 & \multicolumn{1}{r|}{0.81 ± 0.39} & 0.67 ± 0.21 & \multicolumn{1}{r|}{0.89 ± 0.32} & 0.80 ± 0.14 \\ \hline
[11] Marking                & \multicolumn{1}{r|}{1.00 ± 0.07} & 0.91 ± 0.08 & \multicolumn{1}{r|}{0.99 ± 0.10} & 0.80 ± 0.20 & \multicolumn{1}{r|}{0.99 ± 0.07} & 0.90 ± 0.09 \\ \hline
[12] No Action              & \multicolumn{1}{r|}{0.97 ± 0.18} & 0.87 ± 0.11 & \multicolumn{1}{r|}{0.93 ± 0.25} & 0.77 ± 0.20 & \multicolumn{1}{r|}{0.96 ± 0.20} & 0.85 ± 0.13 \\ \hline
[13] Out of Body            & \multicolumn{1}{r|}{0.99 ± 0.09} & 0.98 ± 0.02 & \multicolumn{1}{r|}{0.99 ± 0.11} & 0.96 ± 0.03 & \multicolumn{1}{r|}{0.99 ± 0.09} & 0.97 ± 0.02 \\ \hline
[14] Retraction             & \multicolumn{1}{r|}{0.95 ± 0.22} & 0.78 ± 0.19 & \multicolumn{1}{r|}{0.92 ± 0.26} & 0.66 ± 0.30 & \multicolumn{1}{r|}{0.94 ± 0.24} & 0.77 ± 0.20 \\ \hline
[15] Scope Insertion        & \multicolumn{1}{r|}{1.00 ± 0.07} & 0.95 ± 0.05 & \multicolumn{1}{r|}{0.99 ± 0.07} & 0.94 ± 0.06 & \multicolumn{1}{r|}{1.00 ± 0.06} & 0.96 ± 0.04 \\ \hline
[16] Scope Positioning      & \multicolumn{1}{r|}{0.96 ± 0.19} & 0.87 ± 0.11 & \multicolumn{1}{r|}{0.92 ± 0.27} & 0.73 ± 0.23 & \multicolumn{1}{r|}{0.96 ± 0.19} & 0.87 ± 0.11 \\ \hline
[17] Scope Removal          & \multicolumn{1}{r|}{1.00 ± 0.03} & 1.00 ± 0.00 & \multicolumn{1}{r|}{1.00 ± 0.05} & 0.99 ± 0.01 & \multicolumn{1}{r|}{1.00 ± 0.03} & 1.00 ± 0.00 \\ \hline
[18] Smoke                  & \multicolumn{1}{r|}{1.00 ± 0.07} & 0.87 ± 0.13 & \multicolumn{1}{r|}{0.99 ± 0.10} & 0.71 ± 0.28 & \multicolumn{1}{r|}{0.99 ± 0.08} & 0.82 ± 0.17 \\ \hline
[19] Specimen Removal       & \multicolumn{1}{r|}{1.00 ± 0.03} & 0.97 ± 0.03 & \multicolumn{1}{r|}{1.00 ± 0.04} & 0.95 ± 0.05 & \multicolumn{1}{r|}{1.00 ± 0.02} & 0.99 ± 0.01 \\ \hline
[20] Stitching              & \multicolumn{1}{r|}{0.98 ± 0.15} & 0.93 ± 0.06 & \multicolumn{1}{r|}{0.95 ± 0.21} & 0.85 ± 0.12 & \multicolumn{1}{r|}{0.97 ± 0.16} & 0.92 ± 0.06 \\ \hline
[21] Washout                & \multicolumn{1}{r|}{0.99 ± 0.11} & 0.91 ± 0.08 & \multicolumn{1}{r|}{0.97 ± 0.17} & 0.79 ± 0.20 & \multicolumn{1}{r|}{0.99 ± 0.12} & 0.90 ± 0.09 \\ \hline
\end{tabular}
\end{adjustbox}
\end{table}

\subsection*{Discussion}

\vspace{2mm}
The results confirm that the \texttt{STALNet} model with the ConvNeXt encoder performs well and consistently across phases, tasks, and actions with sufficient training data, as evidenced by low variance in well-represented classes. However, as the number of classes increases---from five phases to 11 tasks to 21 actions---the modelling task becomes more challenging, leading to higher variance and lower performance for less frequent classes. This trend underscores the complexity of handling a larger number of classes and highlights the need for addressing class imbalance. Techniques such as weighted dataloaders and customised loss functions can mitigate these issues, improving the model's robustness and performance across all categories. 

\begin{figure}[!t]
    \centering    \includegraphics[width=1\textwidth]{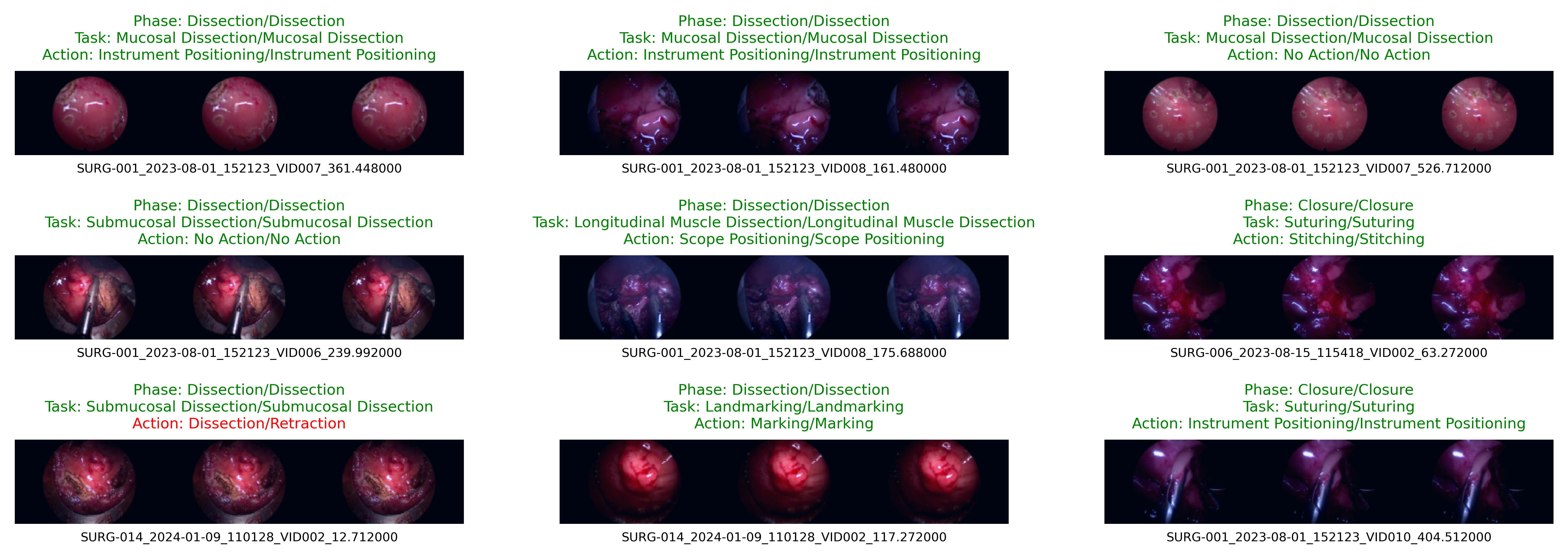}
\caption{\textbf{STALNet: Batch of results for visual inspection:} \textit{This figure illustrates the output of the STALNet model compared to human annotations---the ground truth (GT). Each tile displays the first, middle, and last frames of a video clip, along with predictions and GT for each taxonomy triplet (Phase, Task, Action) at the top. Green font indicates agreement with the GT, while red font indicates disagreement. In this example, there is widespread agreement except for one micro-clip where the model predicted the action "retraction" instead of "dissection" as labeled by the human annotators.}}    \label{fig:results}
\end{figure}

\vspace{2mm}
The results also illustrate the model's superior capabilities in capturing the nuances of surgical workflows. The ROC curves highlights that the Swin V2 encoder outperforms other encoders in terms of accuracy and F1 score. The model's output is visually depicted in an infographic in Figure \ref{fig:results}. This shows the input video clips with predicted and actual taxonomy triplet labels from a batch. This visualisation clearly demonstrates the trends discussed in the performance tables and ROC curves, providing a comprehensive understanding of the model's efficacy in real-world scenarios.

\subsection*{Clinical Utility and Future work}
In this study, the proposed timeline segmentation model has been employed to index a large number of trans-anal endoscopic microsurgery procedures, potentially creating an intuitive front-end platform for surgeons, educators and surgical training committees to analyze surgical videos effectively in a time-effective manner. The methodology described here is generalisable and can be used in any form of surgery where video recording is undertaken. It is possible to create search capabilities of the ESV searching platform, which leverages the timeline segmentation models to efficiently analyse multi-part surgical videos of a single video or across a large library of videos. Surgeons will have the capability of searching within a single video or across their entire ``personal'' surgical video bank using the timeline labels generated by the model. Taking this to the next level, training or NHS governance committees would be able to do this at scale to ensure quality of surgical procedures are maintained at scale and across surgical domains. This personalised searching capability is crucial for improving surgical technique and demonstrating effectiveness in various governance tasks, such as appraisals. 

\vspace{2mm}
Currently, trainees or surgeons do not routinely submit VBAs for appraisal. This is polyfactorial but may be due to video file size, difficulty scrolling through large videos to find ``key steps'' and lack of a reliable standardised process to index videos.

\vspace{2mm}
The model demonstrated here has the potential for clinicians to use STALNet to index their videos so that they can quickly locate video clips of specific intra-operative surgical events from large video datasets. Once the model has been validated in a clinical trial, a future project may focus on video library analysis to identify key behaviours that can be modified to enhance future surgical performance. The timeline also enables comparisons between surgeons based on their surgical behaviour, task efficiency, end-to-end operative progression and intra-operative risk management. Any future clinical feedback system should allow users to reliably and securely filter and sort videos by type, speciality, and hospital, providing a powerful tool for detailed clinical data science analysis. It is important that this is embedded within a SDE to enable secure analysis and feedback. This lays the foundation to create a safe environment where continuous improvement in surgical practices can occur across a large cohort of surgeons. The possibilities offered by this system are vast, empowering clinicians to conduct comprehensive reviews of intra-operative tasks to improve surgical outcomes for patients.

%
%

\section*{Usage Notes}

The dataset described in this study is available, after necessary approvals from the UHB data governance team. This dataset is designed to facilitate the training and evaluation of machine learning models for surgical timeline segmentation based on the proposed taxonomy. Researchers can use several software packages to analyse and process the dataset, with Python being particularly useful for data handling, preprocessing, and model training. Key libraries include FFmpeg and av for video processing and frame extraction, timm for accessing various pre-trained models, PyTorch for deep learning model implementation and training, fastai for simplifying the training process and integrating with PyTorch, nbdev for creating reproducible and literate programs, and Matplotlib for visualising data and model performance. It is recommended to normalise the video frames to standardise the input data. Microclips can be generated based on custom logic using tools like FFmpeg. Additionally, Matplotlib and pandas can be used to analyse data distribution and class imbalance in the dataset.
When integrating or comparing this dataset with others, it is essential to ensure consistent preprocessing steps to maintain uniformity. Utilising common evaluation metrics can help effectively compare performance across different datasets. Considering the temporal nature of surgical workflow data when combining it with static datasets is crucial to preserve contextual information. We will provide detailed information on accessing the data, including criteria for determining eligibility and any limitations on data use. Researchers will need to apply through the specified access control mechanisms to obtain the dataset, ensuring compliance with ethical guidelines and data protection regulations. By following these guidelines and utilising the provided tools and recommendations, researchers can effectively leverage this dataset for advancing the field of surgical timeline segmentation and related applications. For additional resources, code, and tools, please refer to the Code Availability section.

\section*{Code Availability}

The necessary scripts used in the generation and processing of the dataset for this study is available in the GitHub repository at \url{https://github.com/bilalcodehub/evr}. This repository contains all the necessary scripts and tools for working with the dataset. Included are data preprocessing scripts for normalising video frames, generating microclips using FFmpeg, and handling data distribution and class imbalance with pandas and Matplotlib. Additionally, the repository provides model training and evaluation scripts for implementing and training deep learning models using PyTorch and fastai, with configurations for integrating pre-trained models from timm. There are also tools for evaluating model performance using metrics like accuracy, F1 score, and ROC curves, as well as for visualising data and model results with Matplotlib. To ensure reproducibility, nbdev scripts for creating reproducible and literate programs are included.

\vspace{2mm}
The repository also provides detailed documentation on the versions of software used and instructions on how to set up and run the scripts. Specific variables and parameters used to generate, test, and process the current dataset are provided within the scripts, ensuring the study can be replicated accurately. We aim to facilitate the reuse of our dataset and the replication of our study, allowing other researchers to build upon our work in the field of surgical timeline segmentation. For further assistance, please refer to the documentation in the GitHub repository or contact the corresponding authors.

\section*{Acknowledgements}
The authors express their deepest gratitude to the patients who consented to have their procedures recorded; without their invaluable contribution, this project would not have been possible. We also extend our thanks to the theatre and anaesthetic staff at Good Hope Hospital, University Hospitals Birmingham (UK), whose dedication over the past 20 years has established the institution as one of the leading TEMS centers in the UK, demonstrating world-class expertise. Additionally, we are grateful to all staff at University Hospitals Birmingham for fostering an environment that embraces and supports innovation.   

\section*{Author Contributions}

M.B. and S.A. conceived the project idea, S.A., D.B., S.K., N.L., Mo.A. S.B., and A.B. were clinical domain experts and assisted in labelling and validation, M.B., Ma.A., A.H., K.S., I.Q., P.C., Z.K. A.Q., J.Q. and S.A. conducted the experiment and analysis, M.B., Ma.A., A.H., H.V., M.C., A.Q., J.Q. and S.A. analysed the results. M.B., Ma.A. and S.A. drafted the first version of the manuscript. All authors reviewed and helped edit the manuscript. 

\section*{Competing Interests} 
No author declares any competing interest.

\bibliography{sample}

\begin{thebibliography}{10}
\urlstyle{rm}
\expandafter\ifx\csname url\endcsname\relax
  \def\url#1{\texttt{#1}}\fi
\expandafter\ifx\csname urlprefix\endcsname\relax\def\urlprefix{URL }\fi
\expandafter\ifx\csname doiprefix\endcsname\relax\def\doiprefix{DOI: }\fi
\providecommand{\bibinfo}[2]{#2}
\providecommand{\eprint}[2][]{\url{#2}}

\bibitem{weiser2015estimate}
\bibinfo{author}{Weiser, T.~G.} \emph{et~al.}
\newblock \bibinfo{journal}{\bibinfo{title}{Estimate of the global volume of surgery in 2012: an assessment supporting improved health outcomes}}.
\newblock {\emph{\JournalTitle{The Lancet}}} \textbf{\bibinfo{volume}{385}}, \bibinfo{pages}{S11} (\bibinfo{year}{2015}).

\bibitem{nepogodiev2019global}
\bibinfo{author}{Nepogodiev, D.} \emph{et~al.}
\newblock \bibinfo{journal}{\bibinfo{title}{Global burden of postoperative death}}.
\newblock {\emph{\JournalTitle{The Lancet}}} \textbf{\bibinfo{volume}{393}}, \bibinfo{pages}{401} (\bibinfo{year}{2019}).

\bibitem{SDE}
\bibinfo{author}{England, N.}
\newblock \bibinfo{journal}{\bibinfo{title}{Secure data environment}}.
\newblock {\emph{\JournalTitle{www.digital.nhs.uk/services/secure-data-environment-service}}}  (\bibinfo{year}{2021}).

\bibitem{Green}
\bibinfo{author}{NHS, N.~Z.}
\newblock \bibinfo{journal}{\bibinfo{title}{Delivering a net zero nhs}}.
\newblock {\emph{\JournalTitle{https://www.england.nhs.uk/greenernhs/a-net-zero-nhs/}}}  (\bibinfo{year}{2020}).

\bibitem{maier2017surgical}
\bibinfo{author}{Maier-Hein, L.} \emph{et~al.}
\newblock \bibinfo{journal}{\bibinfo{title}{Surgical data science for next-generation interventions}}.
\newblock {\emph{\JournalTitle{Nature Biomedical Engineering}}} \textbf{\bibinfo{volume}{1}}, \bibinfo{pages}{691--696} (\bibinfo{year}{2017}).

\bibitem{reiley2011review}
\bibinfo{author}{Reiley, C.~E.}, \bibinfo{author}{Lin, H.~C.}, \bibinfo{author}{Yuh, D.~D.} \& \bibinfo{author}{Hager, G.~D.}
\newblock \bibinfo{journal}{\bibinfo{title}{Review of methods for objective surgical skill evaluation}}.
\newblock {\emph{\JournalTitle{Surgical endoscopy}}} \textbf{\bibinfo{volume}{25}}, \bibinfo{pages}{356--366} (\bibinfo{year}{2011}).

\bibitem{goodman2021real}
\bibinfo{author}{Goodman, E.~D.} \emph{et~al.}
\newblock \bibinfo{journal}{\bibinfo{title}{A real-time spatiotemporal ai model analyzes skill in open surgical videos}}.
\newblock {\emph{\JournalTitle{arXiv preprint arXiv:2112.07219}}}  (\bibinfo{year}{2021}).

\bibitem{padoy2019machine}
\bibinfo{author}{Padoy, N.}
\newblock \bibinfo{journal}{\bibinfo{title}{Machine and deep learning for workflow recognition during surgery}}.
\newblock {\emph{\JournalTitle{Minimally Invasive Therapy \& Allied Technologies}}} \textbf{\bibinfo{volume}{28}}, \bibinfo{pages}{82--90} (\bibinfo{year}{2019}).

\bibitem{huaulme2020offline}
\bibinfo{author}{Huaulm{\'e}, A.} \emph{et~al.}
\newblock \bibinfo{journal}{\bibinfo{title}{Offline identification of surgical deviations in laparoscopic rectopexy}}.
\newblock {\emph{\JournalTitle{Artificial Intelligence in Medicine}}} \textbf{\bibinfo{volume}{104}}, \bibinfo{pages}{101837} (\bibinfo{year}{2020}).

\bibitem{kadkhodamohammadi2021towards}
\bibinfo{author}{Kadkhodamohammadi, A.} \emph{et~al.}
\newblock \bibinfo{journal}{\bibinfo{title}{Towards video-based surgical workflow understanding in open orthopaedic surgery}}.
\newblock {\emph{\JournalTitle{Computer Methods in Biomechanics and Biomedical Engineering: Imaging \& Visualization}}} \textbf{\bibinfo{volume}{9}}, \bibinfo{pages}{286--293} (\bibinfo{year}{2021}).

\bibitem{holden2014feasibility}
\bibinfo{author}{Holden, M.~S.} \emph{et~al.}
\newblock \bibinfo{journal}{\bibinfo{title}{Feasibility of real-time workflow segmentation for tracked needle interventions}}.
\newblock {\emph{\JournalTitle{IEEE Transactions on Biomedical Engineering}}} \textbf{\bibinfo{volume}{61}}, \bibinfo{pages}{1720--1728} (\bibinfo{year}{2014}).

\bibitem{padoy2012statistical}
\bibinfo{author}{Padoy, N.} \emph{et~al.}
\newblock \bibinfo{journal}{\bibinfo{title}{Statistical modeling and recognition of surgical workflow}}.
\newblock {\emph{\JournalTitle{Medical image analysis}}} \textbf{\bibinfo{volume}{16}}, \bibinfo{pages}{632--641} (\bibinfo{year}{2012}).

\bibitem{lin2005automatic}
\bibinfo{author}{Lin, H.~C.} \emph{et~al.}
\newblock \bibinfo{title}{Automatic detection and segmentation of robot-assisted surgical motions}.
\newblock In \emph{\bibinfo{booktitle}{International conference on medical image computing and computer-assisted intervention}}, \bibinfo{pages}{802--810} (\bibinfo{organization}{Springer}, \bibinfo{year}{2005}).

\bibitem{dergachyova2016automatic}
\bibinfo{author}{Dergachyova, O.}, \bibinfo{author}{Bouget, D.}, \bibinfo{author}{Huaulm{\'e}, A.}, \bibinfo{author}{Morandi, X.} \& \bibinfo{author}{Jannin, P.}
\newblock \bibinfo{journal}{\bibinfo{title}{Automatic data-driven real-time segmentation and recognition of surgical workflow}}.
\newblock {\emph{\JournalTitle{International journal of computer assisted radiology and surgery}}} \textbf{\bibinfo{volume}{11}}, \bibinfo{pages}{1081--1089} (\bibinfo{year}{2016}).

\bibitem{jin2017sv}
\bibinfo{author}{Jin, Y.} \emph{et~al.}
\newblock \bibinfo{journal}{\bibinfo{title}{Sv-rcnet: workflow recognition from surgical videos using recurrent convolutional network}}.
\newblock {\emph{\JournalTitle{IEEE transactions on medical imaging}}} \textbf{\bibinfo{volume}{37}}, \bibinfo{pages}{1114--1126} (\bibinfo{year}{2017}).

\bibitem{blum2010modeling}
\bibinfo{author}{Blum, T.}, \bibinfo{author}{Feu{\ss}ner, H.} \& \bibinfo{author}{Navab, N.}
\newblock \bibinfo{title}{Modeling and segmentation of surgical workflow from laparoscopic video}.
\newblock In \emph{\bibinfo{booktitle}{Medical Image Computing and Computer-Assisted Intervention--MICCAI 2010: 13th International Conference, Beijing, China, September 20-24, 2010, Proceedings, Part III 13}}, \bibinfo{pages}{400--407} (\bibinfo{organization}{Springer}, \bibinfo{year}{2010}).

\bibitem{twinanda2016endonet}
\bibinfo{author}{Twinanda, A.~P.} \emph{et~al.}
\newblock \bibinfo{journal}{\bibinfo{title}{{EndoNet}: a deep architecture for recognition tasks on laparoscopic videos}}.
\newblock {\emph{\JournalTitle{IEEE transactions on medical imaging}}} \textbf{\bibinfo{volume}{36}}, \bibinfo{pages}{86--97} (\bibinfo{year}{2016}).

\bibitem{ramesh2021multi}
\bibinfo{author}{Ramesh, S.} \emph{et~al.}
\newblock \bibinfo{journal}{\bibinfo{title}{Multi-task temporal convolutional networks for joint recognition of surgical phases and steps in gastric bypass procedures}}.
\newblock {\emph{\JournalTitle{International journal of computer assisted radiology and surgery}}} \textbf{\bibinfo{volume}{16}}, \bibinfo{pages}{1111--1119} (\bibinfo{year}{2021}).

\bibitem{gao2021trans}
\bibinfo{author}{Gao, X.}, \bibinfo{author}{Jin, Y.}, \bibinfo{author}{Long, Y.}, \bibinfo{author}{Dou, Q.} \& \bibinfo{author}{Heng, P.-A.}
\newblock \bibinfo{title}{{Trans-SVNet}: Accurate phase recognition from surgical videos via hybrid embedding aggregation transformer}.
\newblock In \emph{\bibinfo{booktitle}{Medical Image Computing and Computer Assisted Intervention--MICCAI 2021: 24th International Conference, Strasbourg, France, September 27--October 1, 2021, Proceedings, Part IV 24}}, \bibinfo{pages}{593--603} (\bibinfo{organization}{Springer}, \bibinfo{year}{2021}).

\bibitem{funke2023tunes}
\bibinfo{author}{Funke, I.}, \bibinfo{author}{Rivoir, D.}, \bibinfo{author}{Krell, S.} \& \bibinfo{author}{Speidel, S.}
\newblock \bibinfo{journal}{\bibinfo{title}{{TUNeS: A Temporal U-Net with Self-Attention for Video-based Surgical Phase Recognition}}}.
\newblock {\emph{\JournalTitle{arXiv preprint arXiv:2307.09997}}}  (\bibinfo{year}{2023}).

\bibitem{huber2020video}
\bibinfo{author}{Huber, M.}
\newblock \bibinfo{title}{Video-based content analysis}.
\newblock In \bibinfo{editor}{Campbell, A.~G.}, \bibinfo{editor}{Hong, L.}, \bibinfo{editor}{Meinel, F.} \& \bibinfo{editor}{Zallio, M.} (eds.) \emph{\bibinfo{booktitle}{Handbook of Research on Multimodal Human Computer Interaction and Pervasive Services}}, \url{10.4324/9780429316647-5} (\bibinfo{publisher}{CRC Press}, \bibinfo{year}{2020}).

\bibitem{liu2009encyclopedia}
\bibinfo{author}{Liu, L.} \& \bibinfo{author}{{\"O}zsu, M.~T.}
\newblock \emph{\bibinfo{title}{Encyclopedia of database systems}}, vol.~\bibinfo{volume}{6} (\bibinfo{publisher}{Springer New York}, \bibinfo{year}{2009}).

\bibitem{feldman2020sages}
\bibinfo{author}{Feldman, L.~S.} \emph{et~al.}
\newblock \bibinfo{journal}{\bibinfo{title}{{SAGES Video-Based Assessment (VBA) program}: a vision for life-long learning for surgeons}}.
\newblock {\emph{\JournalTitle{Surgical endoscopy}}} \textbf{\bibinfo{volume}{34}}, \bibinfo{pages}{3285--3288} (\bibinfo{year}{2020}).

\bibitem{vercauteren2019cai4cai}
\bibinfo{author}{Vercauteren, T.}, \bibinfo{author}{Unberath, M.}, \bibinfo{author}{Padoy, N.} \& \bibinfo{author}{Navab, N.}
\newblock \bibinfo{journal}{\bibinfo{title}{{CAI4CAI}: the rise of contextual artificial intelligence in computer-assisted interventions}}.
\newblock {\emph{\JournalTitle{Proceedings of the IEEE}}} \textbf{\bibinfo{volume}{108}}, \bibinfo{pages}{198--214} (\bibinfo{year}{2019}).

\bibitem{nwoye2020recognition}
\bibinfo{author}{Nwoye, C.~I.} \emph{et~al.}
\newblock \bibinfo{title}{Recognition of instrument-tissue interactions in endoscopic videos via action triplets}.
\newblock In \emph{\bibinfo{booktitle}{Medical Image Computing and Computer Assisted Intervention--MICCAI 2020: 23rd International Conference, Lima, Peru, October 4--8, 2020, Proceedings, Part III 23}}, \bibinfo{pages}{364--374} (\bibinfo{organization}{Springer}, \bibinfo{year}{2020}).

\bibitem{mascagni2022computer}
\bibinfo{author}{Mascagni, P.} \emph{et~al.}
\newblock \bibinfo{journal}{\bibinfo{title}{Computer vision in surgery: from potential to clinical value}}.
\newblock {\emph{\JournalTitle{npj Digital Medicine}}} \textbf{\bibinfo{volume}{5}}, \bibinfo{pages}{163} (\bibinfo{year}{2022}).

\bibitem{lewandrowski2020regional}
\bibinfo{author}{Lewandrowski, K.-U.} \emph{et~al.}
\newblock \bibinfo{journal}{\bibinfo{title}{Regional variations in acceptance, and utilization of minimally invasive spinal surgery techniques among spine surgeons: results of a global survey}}.
\newblock {\emph{\JournalTitle{Journal of Spine Surgery}}} \textbf{\bibinfo{volume}{6}}, \bibinfo{pages}{S260} (\bibinfo{year}{2020}).

\bibitem{richards2015national}
\bibinfo{author}{Richards, M.~K.} \emph{et~al.}
\newblock \bibinfo{journal}{\bibinfo{title}{A national review of the frequency of minimally invasive surgery among general surgery residents: assessment of {ACGME} case logs during 2 decades of general surgery resident training}}.
\newblock {\emph{\JournalTitle{JAMA surgery}}} \textbf{\bibinfo{volume}{150}}, \bibinfo{pages}{169--172} (\bibinfo{year}{2015}).

\bibitem{paysan2021self}
\bibinfo{author}{Paysan, D.}, \bibinfo{author}{Haug, L.}, \bibinfo{author}{Bajka, M.}, \bibinfo{author}{Oelhafen, M.} \& \bibinfo{author}{Buhmann, J.~M.}
\newblock \bibinfo{journal}{\bibinfo{title}{Self-supervised representation learning for surgical activity recognition}}.
\newblock {\emph{\JournalTitle{International Journal of Computer Assisted Radiology and Surgery}}} \textbf{\bibinfo{volume}{16}}, \bibinfo{pages}{2037--2044} (\bibinfo{year}{2021}).

\bibitem{valderrama2022towards}
\bibinfo{author}{Valderrama, N.} \emph{et~al.}
\newblock \bibinfo{title}{Towards holistic surgical scene understanding}.
\newblock In \emph{\bibinfo{booktitle}{International conference on medical image computing and computer-assisted intervention}}, \bibinfo{pages}{442--452} (\bibinfo{organization}{Springer}, \bibinfo{year}{2022}).

\bibitem{ayobi2024pixel}
\bibinfo{author}{Ayobi, N.} \emph{et~al.}
\newblock \bibinfo{journal}{\bibinfo{title}{Pixel-wise recognition for holistic surgical scene understanding}}.
\newblock {\emph{\JournalTitle{arXiv preprint arXiv:2401.11174}}}  (\bibinfo{year}{2024}).

\bibitem{BACH202192}
\bibinfo{author}{Bach, S.~P.} \emph{et~al.}
\newblock \bibinfo{journal}{\bibinfo{title}{Radical surgery versus organ preservation via short-course radiotherapy followed by transanal endoscopic microsurgery for early-stage rectal cancer (trec): a randomised, open-label feasibility study}}.
\newblock {\emph{\JournalTitle{The Lancet Gastroenterology \& Hepatology}}} \textbf{\bibinfo{volume}{6}}, \bibinfo{pages}{92--105} (\bibinfo{year}{2021}).

\bibitem{Gurevych2013-uo}
\bibinfo{author}{Gurevych, I.}, \bibinfo{author}{De~Castilho, R.~E.} \& \bibinfo{author}{Biemann, C.}
\newblock \bibinfo{journal}{\bibinfo{title}{Webanno: A flexible, web-based and visually supported system for distributed annotations}}.
\newblock {\emph{\JournalTitle{51st Annual Meeting...}}}  (\bibinfo{year}{2013}).

\bibitem{liu2022convnet}
\bibinfo{author}{Liu, Z.} \emph{et~al.}
\newblock \bibinfo{title}{A convnet for the 2020s}.
\newblock In \emph{\bibinfo{booktitle}{Proceedings of the IEEE/CVF conference on computer vision and pattern recognition}}, \bibinfo{pages}{11976--11986} (\bibinfo{year}{2022}).

\bibitem{liu2022swin}
\bibinfo{author}{Liu, Z.} \emph{et~al.}
\newblock \bibinfo{title}{Swin transformer v2: Scaling up capacity and resolution}.
\newblock In \emph{\bibinfo{booktitle}{Proceedings of the IEEE/CVF conference on computer vision and pattern recognition}}, \bibinfo{pages}{12009--12019} (\bibinfo{year}{2022}).

\bibitem{steiner2021train}
\bibinfo{author}{Steiner, A.} \emph{et~al.}
\newblock \bibinfo{journal}{\bibinfo{title}{How to train your vit? data, augmentation, and regularization in vision transformers}}.
\newblock {\emph{\JournalTitle{arXiv preprint arXiv:2106.10270}}}  (\bibinfo{year}{2021}).

\bibitem{caron2021emerging}
\bibinfo{author}{Caron, M.} \emph{et~al.}
\newblock \bibinfo{title}{Emerging properties in self-supervised vision transformers}.
\newblock In \emph{\bibinfo{booktitle}{Proceedings of the IEEE/CVF international conference on computer vision}}, \bibinfo{pages}{9650--9660} (\bibinfo{year}{2021}).

\bibitem{howard2020fastai}
\bibinfo{author}{Howard, J.} \& \bibinfo{author}{Gugger, S.}
\newblock \bibinfo{journal}{\bibinfo{title}{Fastai: a layered api for deep learning}}.
\newblock {\emph{\JournalTitle{Information}}} \textbf{\bibinfo{volume}{11}}, \bibinfo{pages}{108} (\bibinfo{year}{2020}).

\end{thebibliography}

\end{document}